# Can an LLM Repair a Language Model in Weight Space Without Gradient Training?

1. Bharath Kumar Bolla[1*], Raghavender Surya Upadhayula[1**], Vishnu Surya Reddy Nandi[2], Ramesh Nandipalli[1]

[1] Independent Researcher, Hyderabad, India
[2] Amazon, Seattle, USA
* Email: bolla111@gmail.com
**Email: upadhyayula.surya@gmail.com

**Abstract**

Although an external language model can select and test repairs to another model's weights, behavior can also be recovered through fixed numerical edits. We assessed the value of interactive selection in a controlled study of Transformer repair. After inspecting the predictions, activation summaries, and weight statistics, the editor selects projection recipes to apply to the target without gradient updates. We analyzed 20 recorded sessions involving five masked and causal language models, comparing the selected edits with fixed projections, scripted and matched-budget black-box search, and low-rank adaptation. Under the core two-layer fault, live selection matches black-box search on SmolLM2 and Qwen2.5. For Pythia, live selection achieves 65.62% completion accuracy, compared with 0% for black-box search and 66.15% for a fixed projection. Under a separate compound fault affecting attention and feed-forward weights, both scripted and LLM-guided selection increase the descriptive five-checkpoint mean from 24.89% to 71.25% using the same operator family. On unmodified checkpoints, scripted projection yields small gains or losses on a separate set of framed questions. The experiments identified cases where feedback changed an edit decision, operator coverage accounted for recovery, or supervised adaptation performed better. Combined with measurements of parameter restoration and preservation, these comparisons establish the scope of controller-guided repair for the tested faults and completion tasks.

## 1 Introduction

A change to a small part of a transformer can disrupt its completions. A repair that restores these completions may still change unrelated outputs or leave the weights far from their original values. Takehi et al. demonstrated that an external language model can inspect and alter a target network's weights using tools [1]. This demonstration raises a comparative question. When a controller proposes a successful edit, how much of the gain comes from its choice of edit, and how much comes from the numerical operator available to any controller?

Prior model-editing studies have made this distinction consequential. ROME uses causal tracing and a rank-one weight update to change factual associations, with separate tests of edit specificity and generalization [2]. Hase et al. find that localization scores can be weak predictors of where an edit will work [3]. Rosati et al. show that short-answer editing metrics capture different behavior from long-form evaluation [4]. We therefore measure completion recovery, distance to the original weights, and preservation on separate probes. Each quantity answers a different question about what the repair changed.

We start from a pretrained checkpoint, introduce a documented fault, and provide the external LLM with a bounded interface for choosing an edit. The editor observes diagnostic outputs and numerical summaries. The runtime applies the requested projection to the target's weight matrices. A fixed rule, a scripted selector, and a black-box optimizer provide comparison policies. We also compare direct repair with low-rank adaptation (LoRA) [5] and evaluate training from repaired weights. The original weights and known perturbation allow us to assess restoration directly.

The main benchmark uses counterbalanced adjective completions, with held-out tests for new target entities and for a syntax-and-distractor shift around the same test entities. Five compact checkpoints span masked-token encoders and causal decoders, whose scoring rules are reported separately. We first examine a localized MLP fault, then vary its support, rank, and layer coverage. Further experiments test compound faults and failures induced by training or simulated quantization. A separate exploratory study starts from unmodified checkpoints. These settings establish which comparisons generalize across the tested faults and which depend on the form of damage.

## 1.1 Research questions

**RQ1.** When does development feedback improve the LLM's first edit, and how does the selected edit compare with fixed projection, scripted search, and black-box search under the specified evaluation budgets?

**RQ2.** How do fault support, rank, layer coverage, and available repair operators affect completion recovery and distance to the original weights?

**RQ3.** When does repair before low-rank adaptation improve held-out completions, and what preservation costs accompany these gains?

## 1.2 Contributions

**Controlled recovery benchmarks were established.** We evaluated five checkpoints with known injected faults, a clean weight reference, and separate measurements for task recovery, parameter restoration, and preservation.

**Controller comparison.** We report 20 recorded sessions, including three matched core seed comparisons and five single-seed fault variants, alongside fixed projections, scripted controls, black-box search, and an oracle inverse.

**Operator and adaptation comparisons were also performed.** We report compound-fault operator ablations, repair-before-training sweeps, and boundary cases for training- and quantization-induced failures. The compound fault comparison matches the candidate sets and development budgets, making the observed ties directly interpretable.

## 1.3 Related work

### 1.3.1 Model editing and evaluation

MEND learns to transform task gradients into local edits [6]. ROME and MEMIT directly modify weights to change factual associations, with MEMIT extending the approach to many edits [2, 7]. Our experiment targets the restoration of prior task behavior under known damage. We retain a clean checkpoint as an audit reference and compare multiple selection policies over explicit repair operators. The preservation probes are limited in length and coverage, a distinction motivated by evidence that short-form and long-form editing evaluations diverge [4].

### 1.3.2 Weight space programming

Takehi et al. described direct weight inspection and modification by language models [1]. Subsequent preprints have studied heterogeneous editing trajectories and uneven transfer [8], reusable weight updates driven by experimental feedback [9], and standardized weight-reading, repair, and construction tasks [10]. Our narrower setting exposes a controlled perturbation, a reference state, and paired controller comparisons. The external studies used different edit spaces and budgets, so their scores are contextual evidence rather than direct baselines.

# 2 Controlled recovery setting

## 2.1 Targets and completion tasks

*Figure 1 summarizes the permitted observations, candidate selection, and separate measurements used to evaluate the repair.*

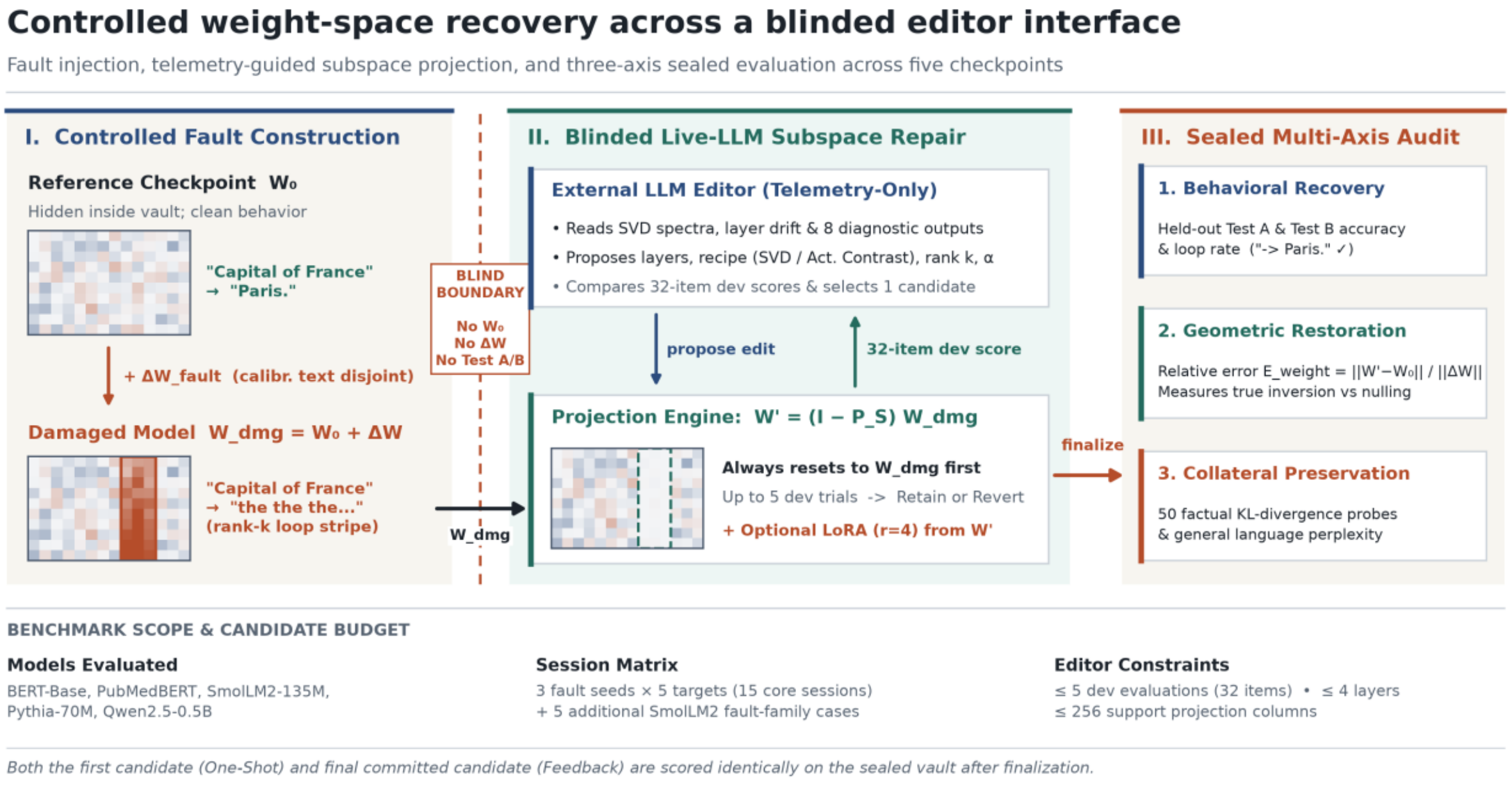

*Figure 1. Overview of the recovery experiment. A known perturbation creates a damaged target. Live LLM and non-LLM procedures receive the damaged state and their permitted observations. Evaluation separates task behavior, geometric restoration, and preservation. Optional adaptation starts from the selected repaired state.*

The target checkpoints were BERT-Base-110M [11], PubMedBERT-110M [12], SmolLM2-135M-Instruct [13], Pythia-70M-Deduped [14], and Qwen2.5-0.5B-Instruct [15]. The encoders predict a single masked token. The decoders generate continuations. The core results retain separate scoring protocols. When the extensions show a pooled mean, it is a descriptive average of checkpoint-specific scores under these protocols.

The controlled task covered eight adjective pairs: cold and hot, low and high, small and large, slow and fast, light and heavy, dark and bright, dry and wet, and sour and sweet. A prompt contained four demonstrations and a query with a distracting first clause followed by a target entity. The answer was one adjective. This was a narrow completion probe, not a comprehensive reasoning benchmark.

For example, two queries can have the identical prefix and first clause, "While fresh snow is white rather than hot," while one ends with a cold target entity and the other with a hot target entity. A function reading only the shared first clause must give the same answer to both and therefore cannot be correct on more than one of them. This construction applies throughout each split. The best possible deterministic lookup using only the complete first clause has exactly 50% accuracy. This is a guarantee about this restricted information, not proof that every conceivable shortcut has been eliminated.

The source produces four items per entity group. This is a paired counterbalancing design, not a complete eight-cell factorial design. There are 128 training-pool items in 32 groups, 32 validation items in eight groups, and 64 items in 16 groups in each test. Prompt strings are disjoint across splits. Training, validation, and Test A use disjoint target entity strings but reuse distractors. Test B reuses Test A's 64 target entities while changing both query syntax and distractors. Consequently, Test B measures a paired syntax-and-distractor shift, not a second test of unseen target entities.

For each seed, adaptation used 32 training examples, one four-item group from each domain. Eight diagnostic examples were taken from the selected groups of two domains. Therefore, the diagnostic set had eight items, with no example from every domain. The editor does not receive the target labels through the diagnostic endpoint. The development evaluation used all 32 labeled validation items.

## 2.2 Fault construction and original weight reference

Let $W_{0,\ell} \in \mathbb{R}^{d_o \times d_i}$ be an MLP output matrix in layer $\ell$. For intermediate activation $h$, its contribution is $W_{0,\ell}h + b$. The injector constructs

$$W_{d,\ell} = W_{0,\ell} + D_\ell.$$

The primary condition changes 64 columns in each of two layers, with a rank-one output direction. The selected zero-based layer pairs are $(7,10)$ for both encoders, $(18,22)$ for SmolLM2, $(2,4)$ for Pythia, and $(16,20)$ for Qwen. These are controlled design choices retained by the host. Ordinary editing responses do not report the true layer pair. The setup metadata and limitations of the information boundary are described below.

An independent calibration prompt identifies highly active intermediate coordinates. The injector shuffles a pool of these coordinates and selects a column support. In the readout-aligned family, its principal output direction is a normalized combination of hesitation-token readouts minus 0.75 times the mean task-answer readout. The injector uses this vocabulary; the localizers and repetition detector do not receive the token list or its resulting vector. This tests a deliberately constructed obstruction, with calibration-text separation but task-aware fault construction.

For rank $R$, the perturbation has the form

$$D_\ell = c_\ell \sum_{r=1}^{R} \frac{1}{\sqrt{R}} v_r s_{\ell,r}^\top,$$

where the $v_r$ are orthonormal output directions and each signed row $s_{\ell,r}^\top$ is supported on the selected columns. The scale uses the median column norm, the model-specific injection strength, $\sqrt{64/K}$ for support size $K$, and $\sqrt{2/L_f}$ for $L_f$ affected layers. The random-direction family removes the mean answer direction from a random vector; it is not orthogonal to every vocabulary readout.

## 2.3 Projection as a shared numerical operation

Let $U_\ell$ contain orthonormal directions in the output space of an MLP matrix, and let $M_\ell$ be a diagonal mask selecting its input columns. The principal repair operation is

$$W_{r,\ell} = W_{d,\ell} - \alpha U_\ell U_\ell^\top W_{d,\ell} M_\ell.$$

The term $U_\ell^\top W_{d,\ell}$ measures how much each column points along the chosen directions. Multiplying by $U_\ell$ reconstructs that component, and $M_\ell$ restricts subtraction to selected columns. At $\alpha = 1$, those columns lose their component along the estimated basis. A smaller strength removes only part of that component.

This operation is shared by live and scripted editors. The changes are in how layers, directions, support, rank, and strength are chosen. The live LLM does not need to calculate a singular-value decomposition in natural language; it requests a recipe, inspects the returned evidence, and chooses a proposal. A fixed algorithm can request the same numerical primitive without an LLM.

Projection is not generally the inverse of the injected change. If the original model already uses the same output direction, the edit removes the original component. For example, let an original weight column be $(1,2)$ and let damage add $(3,0)$. Exact subtraction returns $(1,2)$; removing the entire horizontal component returns $(0,2)$. Both may eliminate the obstruction, but only the first exactly restores the column.

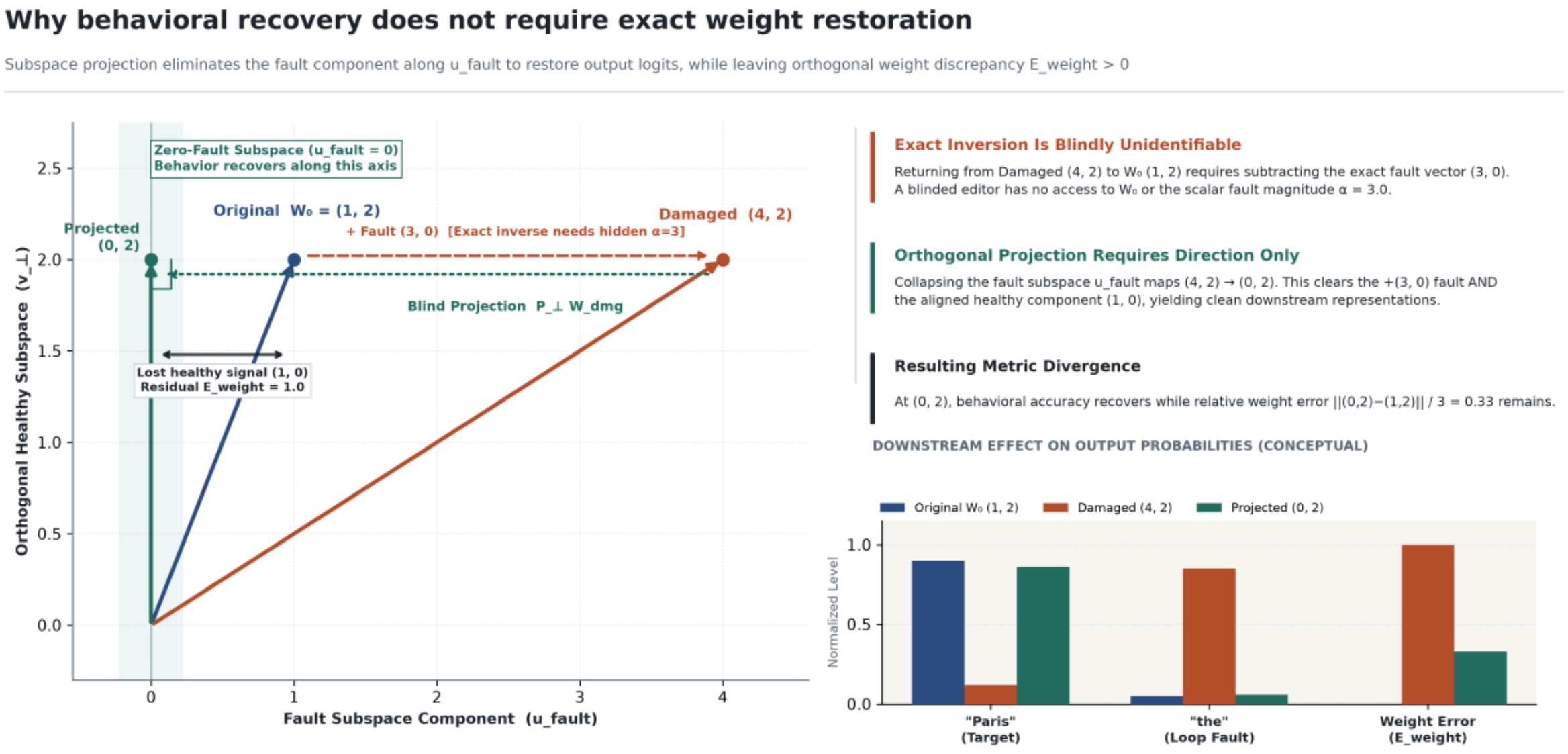

*Figure 2. Projection and exact restoration have different endpoints. The two-dimensional example explains why behavioral recovery can coexist with a residual weight error. Geometric restoration is therefore measured separately from task success.*

# 3 Live LLM editing method and environment

## 3.1 Editor runtime and auditor roles

The editor interacts with a command-line client that sends structured requests to a local HTTP server. That server hosts the damaged target and exposes inspection, proposal, application, development evaluation, reversion, and finalization tools. The editor sees the returned text and numerical summaries. The runtime retains the full tensors and executes the requested matrix operations.

The host also holds privileged audit state: the original MLP matrices, injected changes and supports, and held-out tests. These are not returned by ordinary editing tools. Keeping this information outside routine editor observations allows the evaluator to measure exact restoration without supplying the answer to the repair procedure. The diagnostic endpoint returns model outputs without target answer labels; the development endpoint evaluates against labels on the host and returns aggregate scores.

This is a logical separation of roles and information, not evidence of separate machines or a verified filesystem sandbox. The runtime and audit states coexist in the host implementation. The editor instructions prohibit reading source files and privileged variables, but the saved records do not establish an operating-system access-control boundary. The command-line client exposes both initialization and a final audit action. Initialization returns a fault-family identifier whose descriptive name can encode support or rank; therefore, strict blinding to all fault-family metadata cannot be inferred from the ordinary tool responses alone. The exact injected columns, directions, and original tensors are not part of those responses.

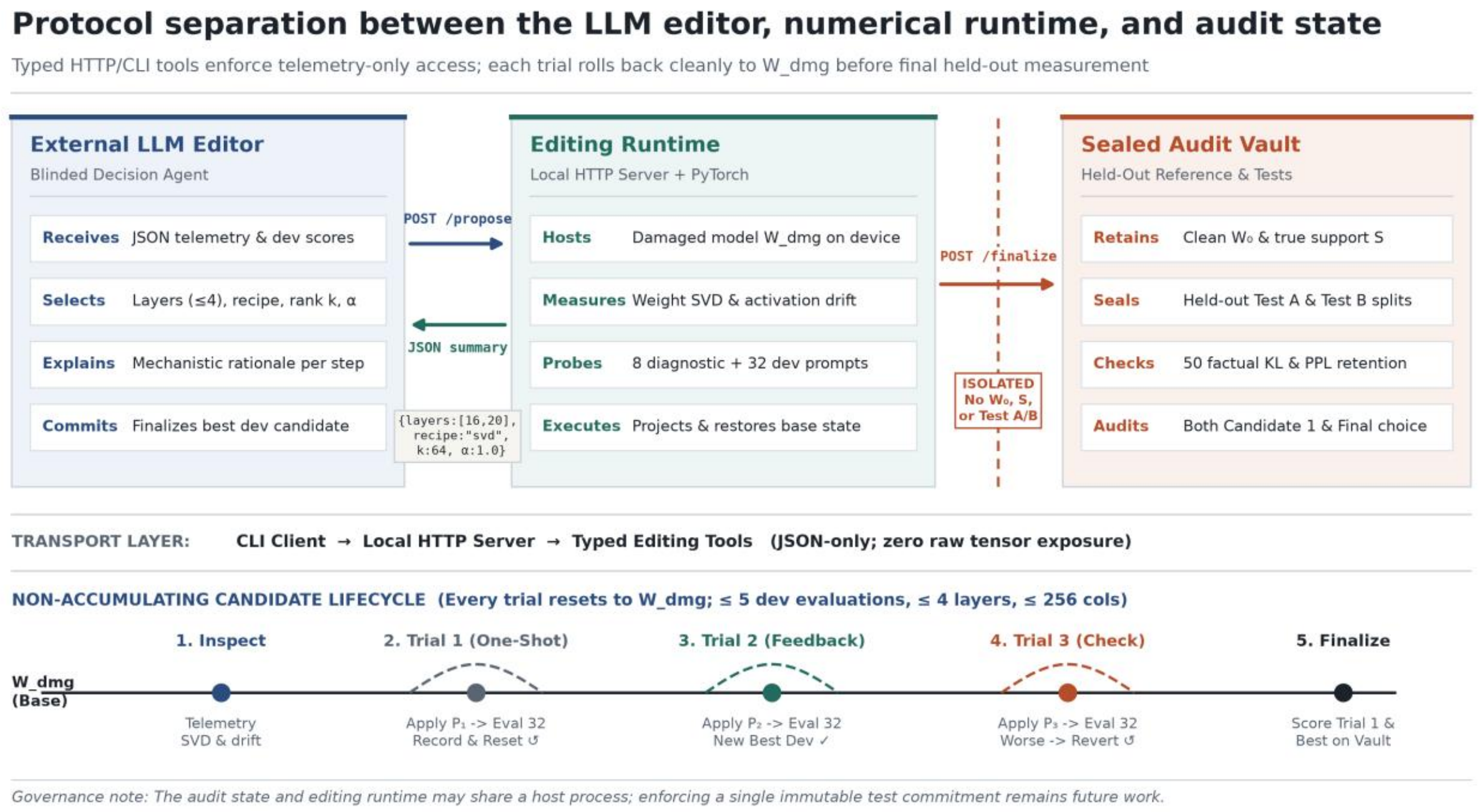

*Figure 3. Logical separation of the external editor, numerical runtime, and audit state. Arrows show requests and returned summaries rather than a security guarantee. Ordinary edit tools omit the original model, injected tensor, and test labels. The local client also exposes setup and audit actions; stronger isolation and immutable post-commit testing require additional enforcement.*

## 3.2 What the tools reveal and change

The inspection tools expose different types of evidence. Model inspection returns the architecture, editable module names, matrix shapes, layer count, data sizes, and evaluation budget. Diagnostic inspection returns short generated continuations or masked-token predictions and repetition or collapse flags on at most eight prompts. Weight inspection returns column-norm statistics, leading singular values, spectral concentration, and anomalous-layer rankings. Activation inspection returns prediction-versus-context anomaly scores, estimated support sizes and ranks, and previews of selected columns. These summaries describe the damaged model rather than a difference from the original model.

A proposal specifies selected MLP layers, a direction recipe, column-selection mode, edit strength, rank mode, and optional context damping. The server returns a proposal identifier, estimated support and rank, predicted update norms, and the number of modified coordinates. The proposal does not change the target. Applying it first restores the saved damaged state and then materializes the requested edit. Consequently, candidate trials start from the same damaged checkpoint rather than accumulating untracked edits.

The development tool evaluates all 32 validation items and returns the completion accuracy, first-token accuracy, candidate-restricted accuracy, repetition rate, and retention-loss probe on eight text sentences. It also reports the evaluation budget remaining. Reversion restores the saved damaged state and checks its state identifier. Finalization materializes the chosen proposal and records its committed state identifier. Returning the damaged state without an edit is an available choice.

The interface permits between one and four selected layers, a column budget up to 256, and explicit edit strengths in [0,1.5]. Fixed-support and median-based adaptive support are available, as are rank-one, heuristic spectral-rank, and explicit-rank modes. These controls constrain the search space and must be included when interpreting the controller's contribution.

### 3.3 Direction recipes and support estimation

The position recipe compares the output of an MLP at the prediction position with its output over the valid preceding context. For diagnostic item $i$ and layer $\ell$,

$$z_{i,\ell} = y_{i,\ell}^{\mathrm{pred}} - \frac{1}{|C_i|}\sum_{t\in C_i} y_{i,\ell,t}, \qquad u_\ell^{(0)} = \frac{\bar{z}_\ell}{\| \bar{z}_\ell \|_2 + \epsilon}.$$

For a masked encoder, the prediction position is the mask. For a causal decoder, it is the final prompt token. Padding and special token positions are excluded from the context. A large difference suggests that a layer contributes unusually strongly at the output position, but it does not by itself identify a causal fault.

Column $j$ receives score $|W_{d,\ell,:,j}^{\mathrm{T}} u_\ell^{(0)}|$. After selecting a support $S_\ell$, the rank-one direction is refined by summing those damaged-weight columns with signs aligned to the initial direction:

$$u_\ell = \frac{\sum_{j\in S_\ell} \mathrm{sign}\,(W_{d,\ell,:,j}^{\mathrm{T}} u_\ell^{(0)}) W_{d,\ell,:,j}}{\left\| \sum_{j\in S_\ell} \mathrm{sign}\,(W_{d,\ell,:,j}^{\mathrm{T}} u_\ell^{(0)}) W_{d,\ell,:,j} \right\|_2 + \epsilon}.$$

The fixed recipe keeps 64 columns. Adaptive support counts projection scores above their median plus 3.5 median absolute deviations, bounded between 16 and 256. Layer ranking compares the projected norm over a reference support with $\sqrt{K}$ times the layer's median column norm. The spectral-rank heuristic uses singular values relative to empirical bulk and leading-value thresholds, capped at four for automatic rank estimation.

The weight-SVD recipe uses no diagnostic forward pass for its direction: it selects high-norm columns and takes the leading left singular vectors of the selected damaged-weight submatrix. The blended recipe aligns the signs of the position and spectral directions and normalizes an 80/20 mixture. Optional context damping reduces the requested projection's component along one context-derived input direction by a factor of 0.25. It is a local damping construction rather than an exact published null-space editor.

Implementation names require care. The confidence-gated localizer computes per-layer strengths from anomaly thresholds, but the live application path applies the explicitly requested proposal strength. It does not automatically multiply by those per-layer gates. The separate scripted gated baseline applies the per-layer strengths and can use a multi-rank basis. Similarly, the rollout diagnostic recipe records failing and passing output counts but currently returns the position-contrast plan without refining its direction from the two groups. We therefore call it rollout diagnostic projection rather than treating it as a separately validated trajectory-contrast estimator.

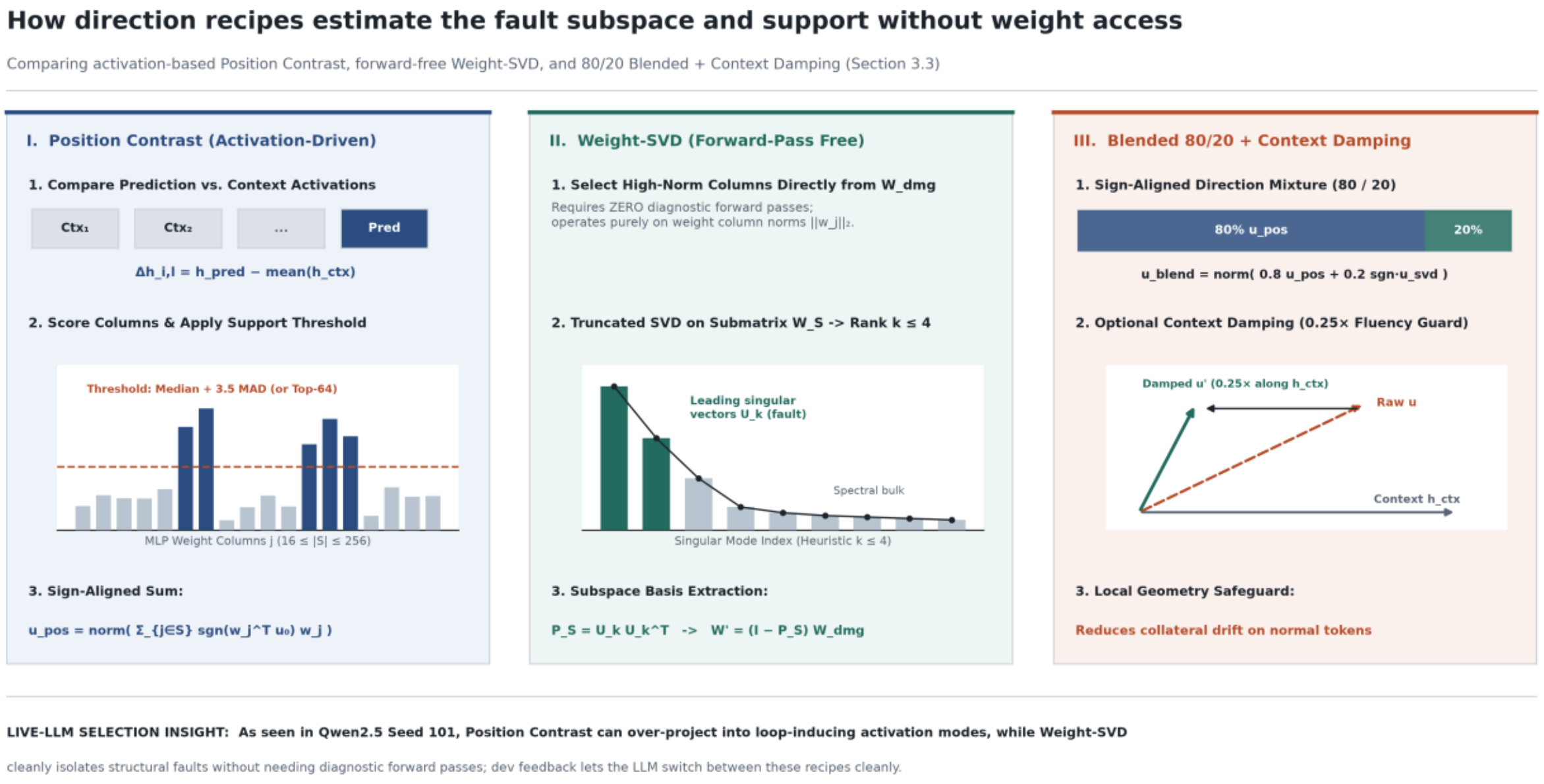

*Figure 4. Conceptual comparison of the three directions and support estimation recipes in Section 3.3. Position Contrast scores weight columns by prediction-versus-context hidden difference; Weight-SVD extracts leading left singular vectors directly from high-norm weight columns without a forward pass; Blended 80/20 combines both directions with optional 0.25× context damping to preserve normal fluency.*

### 3.4 The interactive decision procedure

A live session follows a short decision loop. First, the editor inspects the model structure, outputs, and localization summaries. It then states a hypothesis and proposes a concrete combination of layers, recipe, support, rank, and strength. The runtime applies this proposal to a fresh damaged state. The editor reads the development accuracy, repetition, and retention feedback, then retains the candidate, reverts it, or tests another proposal. After comparing the candidates, it finalizes the choice.

The frozen instructions request a rationale for each call and at least three distinct candidate tests. The allowed development budget is five evaluations. The fifteen core sessions each use three; the five additional SmolLM2 sessions use three or four, for 63 development evaluations overall. Inspection forwards, spectral decompositions, and final test calls are additional work; therefore, five development calls are not a full compute budget.

We report the first candidate from each session and the final feedback-selected candidates. The first candidate is scored retrospectively on the held-out tests after committing to the test. These are two decisions within the same session, not independent randomized one-shot, and feedback experiments. The host measures both against the same damaged reference point. A session’s natural-language rationale is a record of its stated decisions and not an independent measurement of hidden reasoning.

The recorded files retain the fixed prompt, tool arguments, decision summaries, candidate metrics, and state identifiers. All twenty final state identifiers agree with their recorded candidate states, and prompt checksum prefixes are recomputed. The files use a generic editor-role identifier rather than an exact provider model revision; three BERT conversation identifiers are absent. Therefore, complete provider conversations and model revisions are still required for full editor reproducibility.

## 4 Non LLM recovery and adaptation methods

### 4.1 Reference and decoding controls

The original checkpoint establishes the behavior before damage. The damaged checkpoint measures the obstruction. An oracle inverse uses privileged access to the original matrices, equivalently subtracting the injected change, and provides an exact-restoration reference. It is not an ordinary repair competitor because the other methods do not receive this information. The repetition penalty 1.35 changes the decoding rather than the model weights; it tests whether a decoding intervention alone suppresses the induced failure. On encoders, it is an evaluation control, not an autoregressive-loop repair.

### 4.2 Deterministic weight projections

Position projection uses the shared activation contrast, selects two anomalous layers, and removes a rank-one component on 64 estimated columns. Dense position projection uses the same estimated direction but acts on all columns in those layers. The comparison tests whether preserving untouched columns matters.

Gated spectral projection uses a position localizer with automatic spectral rank and per-layer strengths of zero, one-half, or one according to the anomaly thresholds. It differs from the live proposal path described above. Weight-SVD projection chooses layers and directions from the weight spectra alone. Rollout diagnostic projection collects output-failure counts but uses the position plan in the current implementation. Context damping changes the selected projection using a context-derived input direction. These procedures are deterministic, conditional on the damaged state and diagnostic sample; none requires a language-model controller to choose its operations.

### 4.3 Scripted searches over the editing tools

The scripted single-candidate policy inspects the target and applies a fixed rank-one position projection to its two recommended layers. The five-candidate policy uses a predefined schedule that varies the recipe, support, rank, and layer count, when anomaly evidence suggests distributed damage. It ranks development candidates by completion accuracy minus 0.35 times the repetition rate, breaking ties by candidate-restricted accuracy, and then retention loss.

The scripted recipe selector evaluates three fixed proposals: position projection, weight-SVD projection, and a broader position support of 256 columns. The scripted blended-operator control evaluates two combinations of position and spectral directions, with adaptive rank or context damping. To provide a strictly matched black-box optimization control under the same five-evaluation development budget, we evaluate both random search (five sampled layer pairs and strengths) and a matched-budget Bayesian/UCB black-box optimizer (`28_blackbox_bayesian_search_ctrl`). The Bayesian optimizer constructs the candidate layer pool from weight and activation telemetry, executes a three-point space-filling initial design across recipes, support modes, ranks, and scales, and fits an empirical-Bayes additive factor surrogate with an upper confidence bound (UCB) acquisition for its remaining evaluations.

These controls clarify the empirical question. A successful Live LLM repair demonstrates a useful choice within the supplied operator family. A stronger claim that the LLM makes better choices requires comparisons with scripted searches that have the same observation, candidate, and layer budgets. The present policies have related but not identical schedules and capacities; therefore, we report their actual behavior rather than claiming an exact universal compute match.

## 4.4 Low rank adaptation methods

Ordinary LoRA [5] learns a low-rank update on the damaged target,

$$W_{\text{LoRA}} = W_d + BA, \qquad A \in \mathbb{R}^{r \times d_i}, \quad B \in \mathbb{R}^{d_o \times r}.$$

The baseline uses rank four, small-Gaussian initialization for $A$ and zero initialization for $B$. It learns a compensating change rather than explicitly removing the estimated obstruction. Support-masked LoRA restricts the active input coordinates of $A$ to estimated columns. This uses localization information, so it is a more targeted baseline than unrestricted random initialization. Active-coordinate counts and allocated parameters are distinct from the mathematical degrees of freedom of a low-rank matrix.

The DoRA-style implementation [16] separates row magnitudes from an updated direction and normalizes over the input dimension using a detached norm. The magnitude parameters start at the original row norms. The PiSSA-style initialization [17] decomposes the base matrix into leading singular components and a frozen residual, and then trains the low-rank component. Gradient-SVD initialization uses disjoint singular-vector sets from an initial task gradient with a fixed 0.01 scale. It is a local gradient-informed initialization, not a reference-verified reproduction of every gradient-aligned adapter method.

The LoReFT-style implementation [18] changes a hidden representation rather than directly replacing base MLP weights. Its residual intervention is

$$h \mapsto h + R^{\top}(Wh + b - Rh).$$

It starts as the identity, uses a Cayley parametrization for the orthogonal projection, and excludes the rotation parameters from the weight decay. Training and generation intervene at prediction positions, whereas teacher-forced retention evaluation uses all valid positions. These policies affect interpretation. Because the base MLP matrices are unchanged, their restoration error remains one, even if the behavior improves.

The labels DoRA, PiSSA, and LoReFT denote the implemented families. Row normalization, initialization, and orthogonality are present in the code, but execution of forward, gradient, and merged-weight parity with all official reference implementations has not been established. This qualification applies when ranking the families from these results.

## 4.5 Repair followed by training and initialized hybrids

Sequential repair materializes $W_r$ and trains a fresh adapter on that base:

$$W_{\text{seq}} = W_r + BA.$$

We included position repair followed by rank-three or rank-four LoRA, scripted-search repair followed by rank-four LoRA, and Live LLM repair followed by rank-four LoRA. These distinguish the effect of removing an obstruction from the effect of learning an additional task update. Live direct editing uses no target gradients; its repair-plus-LoRA condition adds ordinary gradient training afterward.

The projection-initialized hybrid leaves the damaged base fixed and initializes an adapter to express the projection at step zero.

$$W_{\text{hybrid}} = W_d + B_0 A_0 \quad \text{initially.}$$

Its adapter components subsequently move during optimization. A rank-one fixed repair plus a rank-three adapter has a total update rank of at most four; a rank-one repair plus rank-four adaptation has a cap of five. Equal rank caps do not imply equal parameter counts or optimization trajectories. Functional similarity at step zero should therefore be checked rather than assumed to imply the same learning outcome.

All primary adaptation comparisons use 32 examples, 25 AdamW steps and learning rates selected from $10^{-4}$, $5 \times 10^{-4}$, $1.5 \times 10^{-3}$ and $5 \times 10^{-3}$. The supervised objective predicts the answer token. The current selector prioritizes first-token validation accuracy and then the 16-candidate classification score, whereas the causal test score also penalizes repetitive continuation. This selection mismatch is material and is examined in the results. Longer-training summaries cover 25, 50, and 100 steps where raw checkpoint aggregates exist.

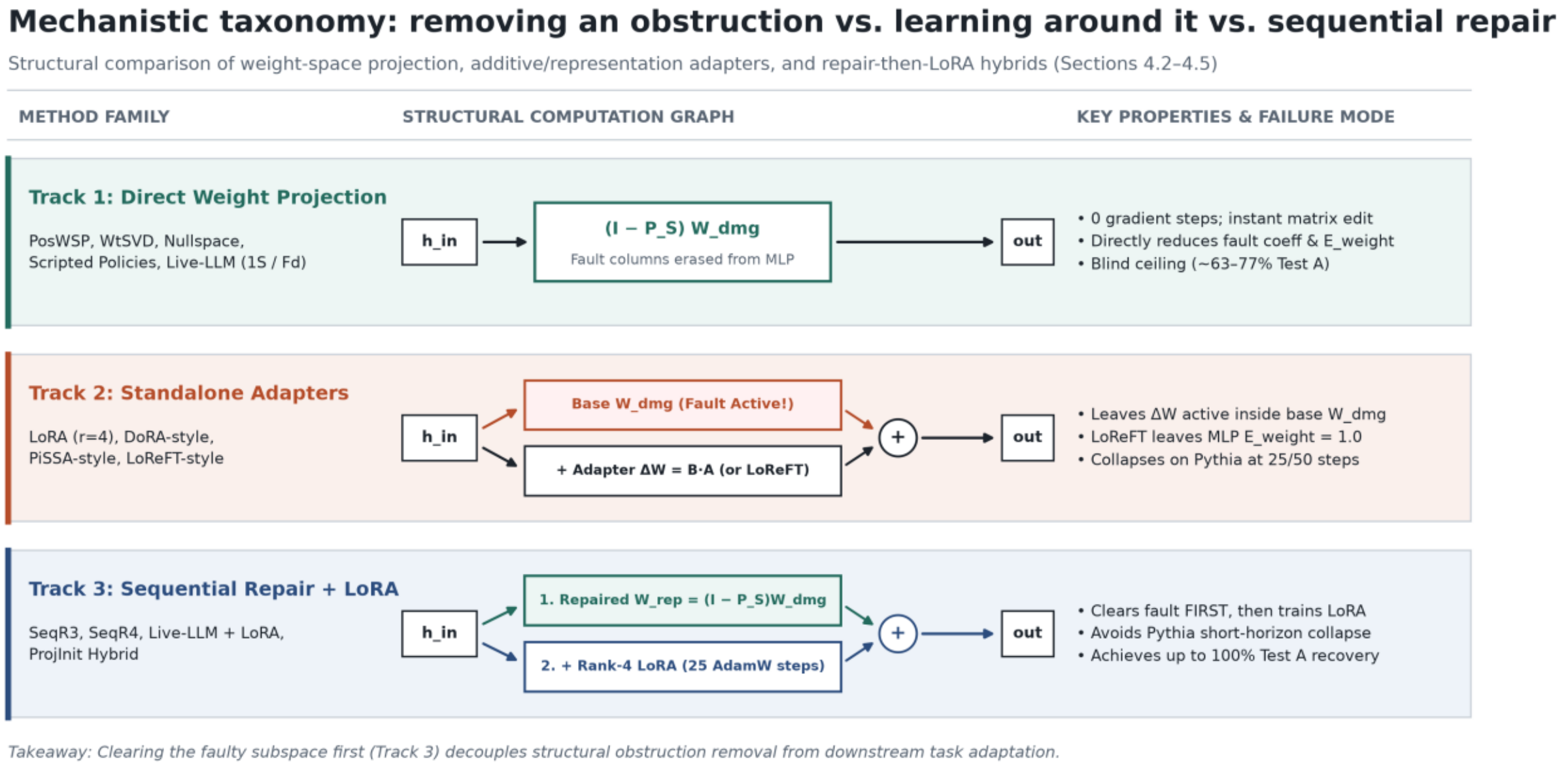

*Figure 5. Mechanistic taxonomy of the recovery and adaptation methods evaluated in Section 4. Direct weight projection (Track 1) removes the estimated fault subspace without gradient updates, standalone adapters (Track 2) leave the fault inside the base weights and risk short-horizon collapse on Pythia-70M, and sequential repair plus LoRA (Track 3) clears the obstruction before fine-tuning.*

# 5 Evaluation protocol and evidence

## 5.1 Behavior restoration and preservation

For the causal models, the evaluator greedily generated up to eight new tokens for every item in both the 64-item tests. It is truncated at the first EOS or decoded newline. An answer is correct when its first whitespace-delimited word matches the expected adjective, and the retained output has no repetition flag. The flag uses token or word repetitions and low diversity, with no hesitation in the word list. It remains a heuristic detector, and the short rollout does not test the long-horizon reasoning loops.

Next-token top-1 accuracy and 16-candidate classification are considered separate metrics. The latter restricts the final choice to the adjective set; domain demonstrations supply further context, so a global 6.25% constant-label baseline is not the only relevant chance comparison. Encoders are evaluated by full-vocabulary masked token top-1 accuracy and cannot exhibit autoregressive looping in this evaluation.

Weight restoration error is

$$E_w = \sqrt{\frac{\sum_\ell \| W_{e,\ell} - W_{0,\ell} \|_F^2}{\sum_\ell \| D_\ell \|_F^2}}.$$

The damaged base has $E_w = 1$ and exact restoration has $E_w = 0$. A representation intervention leaves the base matrices unchanged and therefore has $E_w = 1$ even if outputs improve. An injected-direction coefficient and drift outside the true support provide complementary diagnostics. A reduction from 1 to 0.07 is a 93% reduction in this norm ratio, not a claim that 93% of model capabilities have been restored.

Preservation uses KL divergence from the original output distribution on 50 factual prompts, plus language-modeling loss on 32 technical sentences. Causal loss includes a transition only when both of the adjacent tokens are valid. For encoders, the exponentiated loss masks one selected token per sentence; it is not causal perplexity or full pseudo-perplexity. Both probes are small and do not establish a broad capability retention.

### 5.2 Seed matching and available measurements

The primary live-versus-non-live comparisons used the same three fault seeds, 101, 202, and 303, across all five targets: BERT, PubMedBERT, SmolLM2, Pythia, and Qwen2.5. Every implemented recovery, black-box search, and adaptation family appears in the main results. Additional non-live trials on seeds 404 and 505 appear as expanded five-seed summaries in the appendix.

All five targets—BERT-Base-110M, PubMedBERT-110M, SmolLM2-135M-Instruct, Pythia-70M-Deduped, and Qwen2.5-0.5B-Instruct—are backed by complete raw checkpoint records across all five fault seeds (101, 202, 303, 404, 505), all eight fault-family sweeps (F1–F5), all three PEFT training horizons (25, 50, and 100 steps), and the natural-completion study. Primary tables report matched three-seed (101, 202, 303) comparisons across all five targets, while five-seed non-live summaries are presented in Appendix A.

The automated validation audit verifies the internal arithmetic and raw binary-vector provenance across all five targets. It checks 15,725 saved aggregate fields and 1,920 binary item vectors across the five-seed benchmark and natural-completion study; 100% of binary item vectors exactly reproduce their reported hit counts and means with zero mismatches.

## 6 Results for Live LLM and non LLM methods

### 6.1 Live edits across five targets

*Table 1. Recorded live editing on three fault seeds per target. A scores show percent mean and sample SD; B scores show percent mean. All values come from the recorded session files.*

| Target | First candidate A | Feedback A | Bayesian black-box A | Feedback B | Weight error (Live / Bayes) |
|---|---|---|---|---|---|
| BERT | 63.02 ± 0.90 | 63.02 ± 0.90 | 64.58 ± 1.80 | 50.00 | 0.2709 / 0.6997 |
| PubMed BERT | 72.40 ± 1.81 | 71.88 ± 1.57 | 74.48 ± 2.38 | 53.64 | 0.3689 / 0.6060 |
| SmolLM2 | 74.48 ± 0.90 | 74.48 ± 0.90 | 74.48 ± 0.90 | 75.00 | 0.0678 / 0.0678 |
| Pythia | 66.14 ± 0.91 | 65.62 ± 1.57 | 0.00 ± 0.00 | 14.58 | 0.0734 / 0.4822 |

| Target | First candidate A | Feedback A | Bayesian black-box A | Feedback B | Weight error (Live / Bayes) |
|---|---|---|---|---|---|
| Qwen2.5 | 54.17 ± 38.82 | 76.56 ± 1.56 | 76.56 ± 1.56 | 44.79 | 0.0641 / 0.0641 |

Live editing reduces the injected weight deviation on all five targets. The three causal targets have final mean weight errors close to 0.068, 0.073, and 0.064. On SmolLM2 and Pythia, damage drives task generation close to zero, while both position projection and live editing recover much of the original task behavior. These gains establish that useful behavior can return without gradient training on the damaged target.

For the encoders, the held-out context is more informative than the in-distribution column alone. BERT's matched original Test A is 65.63%, and damaged Test A is 65.10%, but Test B falls from 53.12% to 13.02%. Live feedback brings Test B to 50.00% and weight error to 0.271. PubMedBERT's Test B changes from 64.06% to 45.31% after damage and then to 53.64% after live feedback. The fault affects the encoders even when Test A barely changes; their recovery should not be described as an autoregressive-loop result.

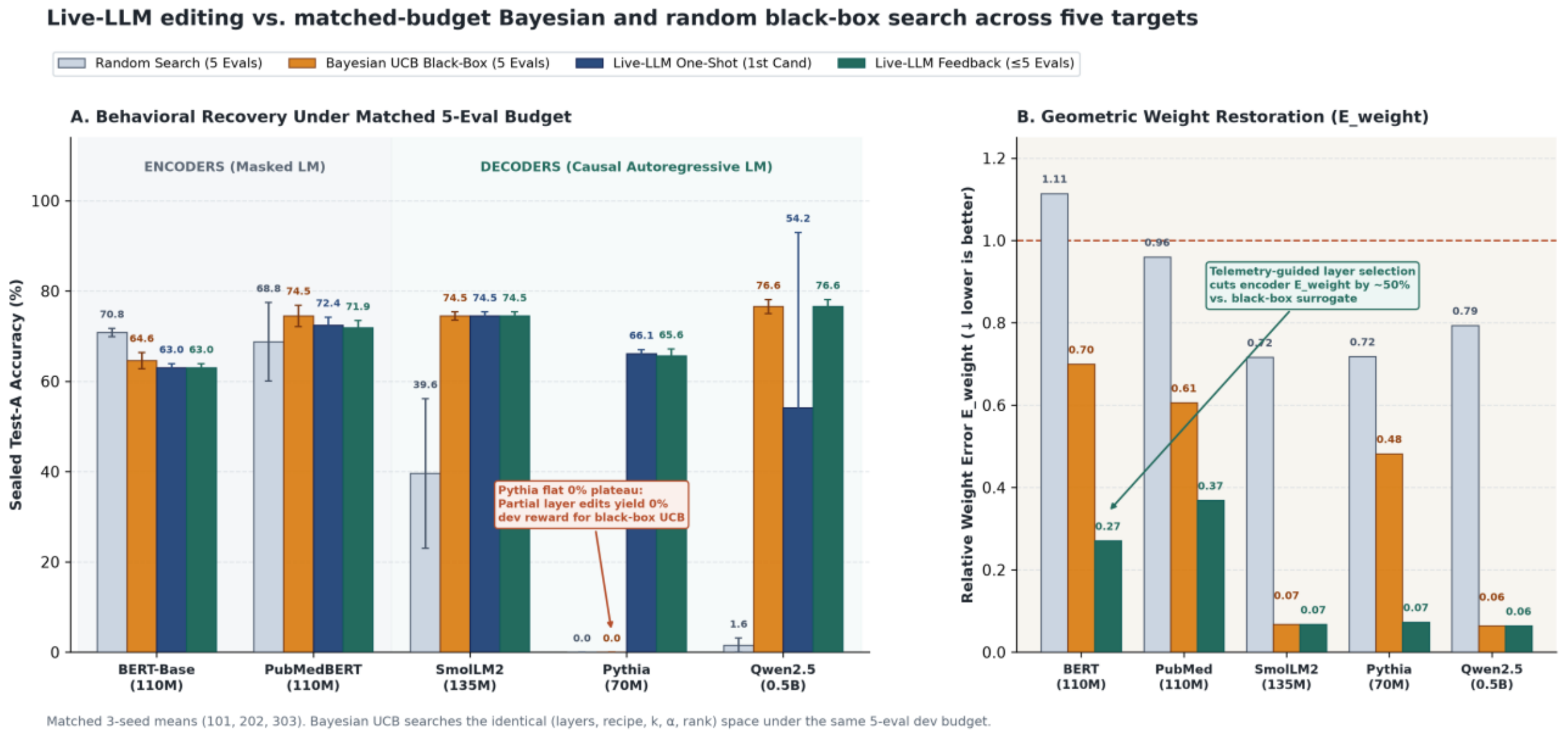

*Figure 6. Matched 3-seed comparison of Live-LLM editing (one-shot and feedback) against matched-budget Bayesian/UCB black-box search (five evaluations) and random search (five evaluations) across all five targets. Left: Sealed Test-A accuracy. Right: Relative weight error, Eweight.*

## 6.2 All direct recovery and scripted control methods

*Table 2. Direct recovery and decoding controls on the same three-fault seeds. Percent mean and sample SD. The oracle uses privileged information, whereas other direct repairs do not receive the original weights. The encoder and decoder scores use different prediction procedures.*

| Method | BERT | PubMed BERT | SmolLM2 | Pythia | Qwen2.5 |
|---|---|---|---|---|---|
| Original checkpoint | 65.62 ± 0.00 | 71.88 ± 0.00 | 75.00 ± 0.00 | 70.31 ± 0.00 | 76.56 ± 0.00 |
| Damaged checkpoint | 65.10 ± 3.93 | 71.87 ± 4.13 | 2.08 ± 3.61 | 0.00 ± 0.00 | 0.00 ± 0.00 |
| Repetition penalty | 28.65 ± 7.71 | 9.38 ± 3.13 | 0.00 ± 0.00 | 0.00 ± 0.00 | 0.00 ± 0.00 |
| Oracle inverse | 65.62 ± 0.00 | 71.88 ± 0.00 | 75.00 ± 0.00 | 70.31 ± 0.00 | 76.56 ± 0.00 |

| Method | BERT | PubMed BERT | SmolLM2 | Pythia | Qwen2.5 |
|---|---|---|---|---|---|
| Position projection | 63.02 ± 0.90 | 72.40 ± 1.81 | 74.48 ± 0.90 | 66.14 ± 0.91 | 54.17 ± 38.82 |
| Gated spectral projection | 58.34 ± 1.81 | 72.40 ± 0.90 | 74.48 ± 0.90 | 66.14 ± 0.91 | 54.17 ± 38.82 |
| Dense position projection | 63.54 ± 0.90 | 72.40 ± 0.90 | 73.96 ± 0.90 | 61.98 ± 4.78 | 75.52 ± 2.38 |
| Rollout diagnostic projection | 63.02 ± 0.90 | 72.40 ± 1.81 | 74.48 ± 0.90 | 66.14 ± 0.91 | 54.17 ± 38.82 |
| Weight SVD projection | 63.02 ± 0.90 | 70.83 ± 0.91 | 74.48 ± 0.90 | 0.00 ± 0.00 | 76.56 ± 1.56 |
| Context damping | 62.50 ± 0.00 | 72.40 ± 2.39 | 73.44 ± 1.56 | 0.00 ± 0.00 | 48.96 ± 42.43 |
| Live LLM first candidate | 63.02 ± 0.90 | 72.40 ± 1.81 | 74.48 ± 0.90 | 66.14 ± 0.91 | 54.17 ± 38.82 |
| Scripted single candidate | 63.02 ± 0.90 | 72.40 ± 1.81 | 74.48 ± 0.90 | 66.14 ± 0.91 | 54.17 ± 38.82 |
| Live LLM feedback | 63.02 ± 0.90 | 71.88 ± 1.57 | 74.48 ± 0.90 | 65.62 ± 1.57 | 76.56 ± 1.56 |
| Scripted five candidates | 65.10 ± 3.25 | 72.40 ± 0.90 | 74.48 ± 0.90 | 66.14 ± 0.91 | 75.52 ± 2.38 |
| Scripted recipe selector | 63.54 ± 0.90 | 71.88 ± 1.57 | 73.96 ± 0.90 | 66.14 ± 0.91 | 75.00 ± 1.56 |
| Scripted blended operator | 61.98 ± 0.90 | 74.48 ± 0.90 | 74.48 ± 0.90 | 4.69 ± 5.42 | 74.48 ± 2.38 |
| Random search (5 evals) | 70.83 ± 0.91 | 68.75 ± 8.70 | 39.59 ± 16.56 | 0.00 ± 0.00 | 1.56 ± 1.56 |
| Bayesian black-box search (5 evals) | 64.58 ± 1.80 | 74.48 ± 2.38 | 74.48 ± 0.90 | 0.00 ± 0.00 | 76.56 ± 1.56 |

The fixed-position projection is a strong core-fault baseline. On matched seeds, it scores 74.48% on SmolLM2 and 66.15% on Pythia, compared with the original scores of 75.00% and 70.31%, respectively. The final live scores were 74.48% and 65.62%. Thus, these core cases show effective repair by both procedures, not a general live-controller advantage. The scripted five-candidate policy also scores 74.48% and 66.15% on these targets. BERT's fixed position, weight-SVD, and final live scores coincide at 63.02% on Test A, even though their geometric and held-out-context outcomes differ.

Pythia's weight-SVD projection scores zero on Test A even though recorded core trials select the planted layer pair. Therefore, correct layer selection is insufficient; direction and support also matter. Dense projection and context damping likewise do not dominate their sparse counterparts. The oracle inverse returns the original behavior, providing a reference for exact restoration. A repetition penalty alone does not recover collapsed causal core cases. Broad random layer search is also weak in this configuration, although its search space and candidate schedule differ from the diagnostic policies.

## 6.3 All adaptation and repair plus training methods

*Table 3. All adaptation methods on the same three fault seeds. Percent mean and sample SD at the 25-step budget. Both scripted and Live LLM repair followed by training are included. The method names denote the local implementations described in the text.*

| Method | BERT | PubMed BERT | SmolLM2 | Pythia | Qwen2.5 |
|---|---|---|---|---|---|
| LoRA | 97.40 ± 2.39 | 83.85 ± 9.15 | 86.98 ± 5.92 | 0.00 ± 0.00 | 89.58 ± 6.31 |
| DoRA style | 97.40 ± 2.39 | 84.89 ± 8.60 | 86.46 ± 6.50 | 0.00 ± 0.00 | 90.10 ± 7.05 |
| PiSSA style | 95.31 ± 2.71 | 86.46 ± 1.80 | 84.37 ± 9.50 | 0.00 ± 0.00 | 97.40 ± 0.90 |
| LoReFT style | 94.27 ± 2.39 | 90.10 ± 1.81 | 70.31 ± 18.42 | 0.00 ± 0.00 | 94.27 ± 2.39 |
| Support masked LoRA | 82.81 ± 8.27 | 83.86 ± 9.55 | 82.82 ± 2.71 | 39.06 ± 9.76 | 92.19 ± 3.13 |

| Method | BERT | PubMed BERT | SmolLM2 | Pythia | Qwen2.5 |
|---|---|---|---|---|---|
| Position repair then LoRA r3 | 95.83 ± 0.91 | 84.38 ± 4.69 | 95.84 ± 1.81 | 37.50 ± 40.87 | 98.44 ± 2.71 |
| Position repair then LoRA r4 | 96.88 ± 0.00 | 84.37 ± 4.13 | 92.71 ± 4.78 | 55.73 ± 38.82 | 100.00 ± 0.00 |
| Projection initialized LoRA | 94.79 ± 2.39 | 89.58 ± 2.39 | 95.31 ± 1.57 | 27.60 ± 25.88 | 99.48 ± 0.90 |
| Gradient SVD initialization | 97.40 ± 0.90 | 89.06 ± 0.00 | 81.77 ± 14.09 | 0.00 ± 0.00 | 84.89 ± 14.18 |
| Live LLM repair then LoRA | 96.36 ± 0.91 | 84.37 ± 4.13 | 95.32 ± 2.71 | 71.35 ± 12.14 | 100.00 ± 0.00 |
| Scripted repair then LoRA | 96.36 ± 0.91 | 86.98 ± 2.38 | 92.71 ± 4.78 | 47.92 ± 40.27 | 100.00 ± 0.00 |

Training can exceed the original task accuracy, but the ranking depends on the target and initialization. On BERT, ordinary LoRA reaches 97.40%, exceeding the 96.35% live-repair-plus-LoRA result. On PubMedBERT, the corresponding means are 83.85% and 84.38%; scripted repair plus LoRA reaches 86.98%. On SmolLM2, live repair then LoRA reaches 95.31%, compared with 86.98% for ordinary LoRA and 92.71% for position repair then rank-four LoRA. On Pythia, ordinary LoRA remains at zero while live repair then LoRA reaches 71.35% at this training budget.

The DoRA-style, PiSSA-style, LoReFT-style, and support-masked and initialized-adapter rows are part of the primary comparison rather than omitted alternatives. They test different routes to compensation: changing magnitudes, choosing informed low-rank factors, modifying representations, or concentrating training on estimated support. Their results should be read with the implementation definitions above. A family name alone does not establish parity with every published reference or guarantee that its default hyperparameters are optimal here.

Across five seeds, position repair and sequential rank-four LoRA both average 65.31%, but their SDs are 2.57 and 30.51 points, respectively. Sequential seed 303 produces only 7/64 valid completions despite 81.25% first-token accuracy because 87.50% of retained outputs are repetitive. The correct first word does not ensure a usable completion. Full-generation development selection and intermediate checkpoints are the most direct next controls.

## 6.4 Held out context across the full method suite

*Table 4. Held-out syntax and distractor shift for every method on the same three fault seeds. Percent mean and sample SD. Target entities are shared with the first test; therefore, this measures a paired context shift rather than an independent entity set.*

| Method | BERT | PubMed BERT | SmolLM2 | Pythia | Qwen2.5 |
|---|---|---|---|---|---|
| Original checkpoint | 53.12 ± 0.00 | 64.06 ± 0.00 | 76.56 ± 0.00 | 14.06 ± 0.00 | 45.31 ± 0.00 |
| Damaged checkpoint | 13.02 ± 3.93 | 45.31 ± 4.14 | 0.00 ± 0.00 | 0.00 ± 0.00 | 0.00 ± 0.00 |
| Repetition penalty | 0.00 ± 0.00 | 3.64 ± 2.39 | 0.00 ± 0.00 | 0.00 ± 0.00 | 0.00 ± 0.00 |
| Oracle inverse | 53.12 ± 0.00 | 64.06 ± 0.00 | 76.56 ± 0.00 | 14.06 ± 0.00 | 45.31 ± 0.00 |
| Position projection | 44.79 ± 3.25 | 54.69 ± 5.42 | 74.48 ± 0.90 | 14.58 ± 2.39 | 29.69 ± 25.72 |
| Gated spectral projection | 50.00 ± 1.56 | 64.58 ± 0.90 | 74.48 ± 0.90 | 14.58 ± 2.39 | 29.69 ± 25.72 |
| Dense position projection | 45.31 ± 2.71 | 54.69 ± 0.00 | 72.40 ± 1.81 | 15.62 ± 1.57 | 43.75 ± 0.00 |
| Rollout diagnostic projection | 44.79 ± 3.25 | 54.69 ± 5.42 | 74.48 ± 0.90 | 14.58 ± 2.39 | 29.69 ± 25.72 |
| Weight SVD projection | 50.00 ± 1.56 | 63.54 ± 5.02 | 75.00 ± 1.56 | 0.00 ± 0.00 | 44.79 ± 0.90 |

| Method | BERT | PubMed BERT | SmolLM2 | Pythia | Qwen2.5 |
|---|---|---|---|---|---|
| Context damping | 47.91 ± 4.51 | 56.77 ± 3.61 | 75.52 ± 2.38 | 0.00 ± 0.00 | 17.71 ± 17.21 |
| LoRA | 82.29 ± 5.92 | 76.56 ± 2.70 | 75.52 ± 11.09 | 0.00 ± 0.00 | 78.65 ± 23.92 |
| DoRA style | 80.73 ± 5.92 | 79.17 ± 5.02 | 76.56 ± 9.76 | 0.00 ± 0.00 | 80.21 ± 19.79 |
| PiSSA style | 75.52 ± 12.73 | 82.29 ± 3.61 | 78.65 ± 11.83 | 0.00 ± 0.00 | 90.62 ± 1.57 |
| LoReFT style | 81.77 ± 3.26 | 79.69 ± 4.14 | 62.50 ± 12.50 | 0.00 ± 0.00 | 84.90 ± 6.51 |
| Support masked LoRA | 65.63 ± 8.27 | 80.73 ± 7.04 | 71.35 ± 5.03 | 26.56 ± 7.16 | 88.54 ± 5.49 |
| Position repair then LoRA r3 | 80.21 ± 11.93 | 80.21 ± 3.25 | 92.19 ± 1.57 | 21.87 ± 29.81 | 83.85 ± 5.91 |
| Position repair then LoRA r4 | 81.25 ± 8.27 | 81.77 ± 4.78 | 91.66 ± 1.81 | 49.48 ± 29.41 | 88.02 ± 8.02 |
| Projection initialized LoRA | 85.42 ± 5.48 | 86.46 ± 2.38 | 94.79 ± 0.90 | 8.33 ± 10.63 | 92.71 ± 3.94 |
| Gradient SVD initialization | 82.29 ± 6.32 | 85.42 ± 3.93 | 54.17 ± 29.91 | 0.00 ± 0.00 | 76.04 ± 21.33 |
| Live LLM first candidate | 44.79 ± 3.25 | 54.69 ± 5.42 | 74.48 ± 0.90 | 14.58 ± 2.39 | 29.69 ± 25.72 |
| Scripted single candidate | 44.79 ± 3.25 | 54.69 ± 5.42 | 74.48 ± 0.90 | 14.58 ± 2.39 | 29.69 ± 25.72 |
| Live LLM feedback | 50.00 ± 1.56 | 53.64 ± 3.61 | 75.00 ± 1.56 | 14.58 ± 2.39 | 44.79 ± 0.90 |
| Scripted five candidates | 48.96 ± 0.90 | 54.16 ± 13.65 | 76.04 ± 2.38 | 14.58 ± 2.39 | 39.06 ± 9.51 |
| Scripted recipe selector | 45.83 ± 7.38 | 53.64 ± 3.61 | 74.48 ± 1.80 | 14.58 ± 2.39 | 38.02 ± 10.05 |
| Scripted blended operator | 48.96 ± 0.90 | 64.06 ± 0.00 | 73.96 ± 2.38 | 0.00 ± 0.00 | 35.94 ± 7.16 |
| Live LLM repair then LoRA | 80.21 ± 6.31 | 83.33 ± 2.39 | 89.06 ± 2.71 | 53.12 ± 14.91 | 84.90 ± 3.25 |
| Scripted repair then LoRA | 80.73 ± 7.71 | 78.65 ± 3.25 | 90.10 ± 2.39 | 33.33 ± 40.30 | 85.42 ± 3.93 |
| Random search (5 evals) | 41.15 ± 6.51 | 42.71 ± 7.04 | 13.54 ± 8.02 | 0.00 ± 0.00 | 0.00 ± 0.00 |
| Bayesian black-box search (5 evals) | 34.90 ± 13.65 | 49.48 ± 7.38 | 75.00 ± 1.56 | 0.00 ± 0.00 | 44.79 ± 0.90 |

Test B checks whether the selected behavior persists when syntax and distractors change around the same target entities. Its complete method table prevents an in-distribution ranking from standing in for recovery. For BERT, a fixed weight-SVD projection reaches 50.00%, while position projection reaches 44.79%; live feedback also reaches 50.00%. PubMedBERT's spectral projection reaches 63.54%, higher than its live-feedback mean of 53.64% in this sample. Thus, a fixed non-LLM recipe can outperform the feedback-selected edit on a held-out metric.

Adaptation often improves these task-specific scores, including on the encoders. On Pythia, live repair followed by LoRA reaches 53.12% on Test B, while the original checkpoint scores 14.06%. This is an adaptation to the completion distribution; it is not evidence that the edited model has recovered a broad capability missing from its original weights. The paired entity construction and narrow answer domains limit this interpretation.

## 6.5 Weight restoration across the full method suite

*Table 5. Mean relative weight error on the same three fault seeds. Zero denotes the exact restoration of the original model, and one is the damaged reference. Adaptation may improve task behavior while moving farther from the original matrices.*

| Method | BERT | PubMed BERT | SmolLM2 | Pythia | Qwen2.5 |
|---|---|---|---|---|---|
| Original checkpoint | 0.000 | 0.000 | 0.000 | 0.000 | 0.000 |

| Method | BERT | PubMed BERT | SmolLM2 | Pythia | Qwen2.5 |
|---|---|---|---|---|---|
| Damaged checkpoint | 1.000 | 1.000 | 1.000 | 1.000 | 1.000 |
| Repetition penalty | 1.000 | 1.000 | 1.000 | 1.000 | 1.000 |
| Oracle inverse | 0.000 | 0.000 | 0.000 | 0.000 | 0.000 |
| Position projection | 0.351 | 0.336 | 0.068 | 0.073 | 0.172 |
| Gated spectral projection | 0.490 | 0.465 | 0.068 | 0.073 | 0.172 |
| Dense position projection | 0.216 | 0.278 | 0.128 | 0.226 | 0.172 |
| Rollout diagnostic projection | 0.351 | 0.336 | 0.068 | 0.073 | 0.172 |
| Weight SVD projection | 0.271 | 0.370 | 0.068 | 0.390 | 0.064 |
| Context damping | 0.353 | 0.333 | 0.086 | 0.127 | 0.203 |
| LoRA | 1.014 | 1.024 | 1.013 | 1.395 | 1.176 |
| DoRA style | 1.012 | 1.024 | 1.010 | 1.333 | 1.159 |
| PiSSA style | 1.101 | 1.173 | 1.018 | 1.341 | 1.202 |
| LoReFT style | 1.000 | 1.000 | 1.000 | 1.000 | 1.000 |
| Support masked LoRA | 1.016 | 1.029 | 1.001 | 1.160 | 1.086 |
| Position repair then LoRA r3 | 0.424 | 0.354 | 0.153 | 0.288 | 0.357 |
| Position repair then LoRA r4 | 0.401 | 0.371 | 0.134 | 0.226 | 0.228 |
| Projection initialized LoRA | 0.623 | 0.406 | 0.307 | 0.378 | 0.298 |
| Gradient SVD initialization | 1.016 | 1.040 | 1.004 | 1.346 | 1.119 |
| Live LLM first candidate | 0.351 | 0.336 | 0.068 | 0.073 | 0.172 |
| Scripted single candidate | 0.351 | 0.336 | 0.068 | 0.073 | 0.172 |
| Live LLM feedback | 0.271 | 0.369 | 0.068 | 0.073 | 0.064 |
| Scripted five candidates | 0.657 | 0.638 | 0.083 | 0.073 | 0.145 |
| Scripted recipe selector | 0.291 | 0.369 | 0.119 | 0.073 | 0.186 |
| Scripted blended operator | 0.402 | 0.561 | 0.103 | 0.292 | 0.181 |
| Live LLM repair then LoRA | 0.334 | 0.402 | 0.159 | 0.253 | 0.110 |
| Scripted repair then LoRA | 0.692 | 0.671 | 0.149 | 0.277 | 0.180 |
| Random search (5 evals) | 1.114 | 0.960 | 0.716 | 0.719 | 0.794 |
| Bayesian black-box search (5 evals) | 0.700 | 0.606 | 0.068 | 0.482 | 0.064 |

Direct projections typically stay much closer to the original MLP matrices than learning a compensating adapter on the damaged base. This geometric advantage is distinct from maximum task accuracy. Exact inverse subtraction has zero restoration error. Representation intervention leaves the measured base matrices unchanged and has an error of one. Low-rank adaptation can improve task behavior while retaining the planted disturbance and adding a new update, producing errors at or above one. Repair followed by adaptation moves away from the repaired state again; therefore, it should be judged jointly on task gain and preservation.

*Table 6. Matched three-seed tasks and preservation measures. Test B is the percentage of correct answers. Loss exp is a causal perplexity for decoders and a selected-mask exponentiated loss for encoders; these are not comparable across architectures. Fact KL is in nats.*

| Model | Method | Test B | Loss exp | Fact KL | Weight error |
|---|---|---|---|---|---|
| BERT | Original | 53.12 | 2.17 | 0.000 | 0.000 |
| BERT | Bayesian black-box search | 34.90 | 2.18 | 0.448 | 0.700 |
| BERT | Live LLM feedback | 50.00 | 2.17 | 0.022 | 0.271 |
| BERT | LoRA | 82.29 | 2.61 | 1.829 | 1.014 |
| BERT | Live LLM repair then LoRA | 80.21 | 2.36 | 1.437 | 0.334 |
| PubMed BERT | Original | 64.06 | 3.95 | 0.000 | 0.000 |
| PubMed BERT | Bayesian black-box search | 49.48 | 4.50 | 0.230 | 0.606 |
| PubMed BERT | Live LLM feedback | 53.64 | 4.14 | 0.088 | 0.369 |
| PubMed BERT | LoRA | 76.56 | 9.25 | 3.023 | 1.024 |
| PubMed BERT | Live LLM repair then LoRA | 83.33 | 5.17 | 1.332 | 0.402 |
| SmolLM2 | Original | 76.56 | 66.40 | 0.000 | 0.000 |
| SmolLM2 | Bayesian black-box search | 75.00 | 66.34 | 0.007 | 0.068 |
| SmolLM2 | Live LLM feedback | 75.00 | 66.34 | 0.007 | 0.068 |
| SmolLM2 | LoRA | 75.52 | 184.29 | 2.226 | 1.013 |
| SmolLM2 | Live LLM repair then LoRA | 89.06 | 67.99 | 0.065 | 0.159 |
| Pythia | Original | 14.06 | 194.28 | 0.000 | 0.000 |
| Pythia | Bayesian black-box search | 0.00 | 1.37e+06 | 5.757 | 0.482 |
| Pythia | Live LLM feedback | 14.58 | 201.66 | 0.112 | 0.073 |
| Pythia | LoRA | 0.00 | 1.37e+08 | 12.405 | 1.395 |
| Pythia | Live LLM repair then LoRA | 53.12 | 342.92 | 4.835 | 0.253 |
| Qwen2.5 | Original | 45.31 | 20.95 | 0.000 | 0.000 |
| Qwen2.5 | Bayesian black-box search | 44.79 | 20.94 | 0.006 | 0.064 |
| Qwen2.5 | Live LLM feedback | 44.79 | 20.93 | 0.006 | 0.064 |
| Qwen2.5 | LoRA | 78.65 | 47.09 | 2.431 | 1.176 |
| Qwen2.5 | Live LLM repair then LoRA | 84.90 | 22.01 | 0.042 | 0.110 |

The small preservation probes show material tradeoffs. Pythia’s causal PPL is 194.28 originally, 201.66 after live repair, and 342.92 after live repair plus LoRA; factual KL changes from zero to 0.112 and 4.835 nats. SmolLM2 changes from PPL 66.40 to 66.34 after live repair and 67.99 after repair plus LoRA. These measurements favor direct repair on preservation in the reported cases, without establishing broad retention guarantees from fifty factual prompts and thirty-two sentences.

## 6.6 How live feedback changes a concrete edit

Qwen seed 101 provides an intuitive example. The first candidate selects layers 16 and 20, 64 columns and rank one, using the confidence-labeled activation recipe. Development completion accuracy is 12.50% with a 46.88% repetition rate. The second candidate keeps the same layers and support but requests the weight-SVD direction. Development accuracy rises to 78.12% with no repetition flags. A third blended candidate fails, and the editor selects the second.

The sealed first-candidate score is 6/64, and its weight error is 0.3855. The selected candidate reaches 49/64 with an error of 0.0634. The change is in direction recipe, not in layer pair. Across Qwen's three sessions, the first-candidate mean accuracy is 54.17%, and the final mean accuracy is 76.56%; the increase is driven by this one session. The other two retain their first candidates. This is a useful repair decision, while a suitably selected scripted spectral recipe could, in principle, make the same numerical edit.

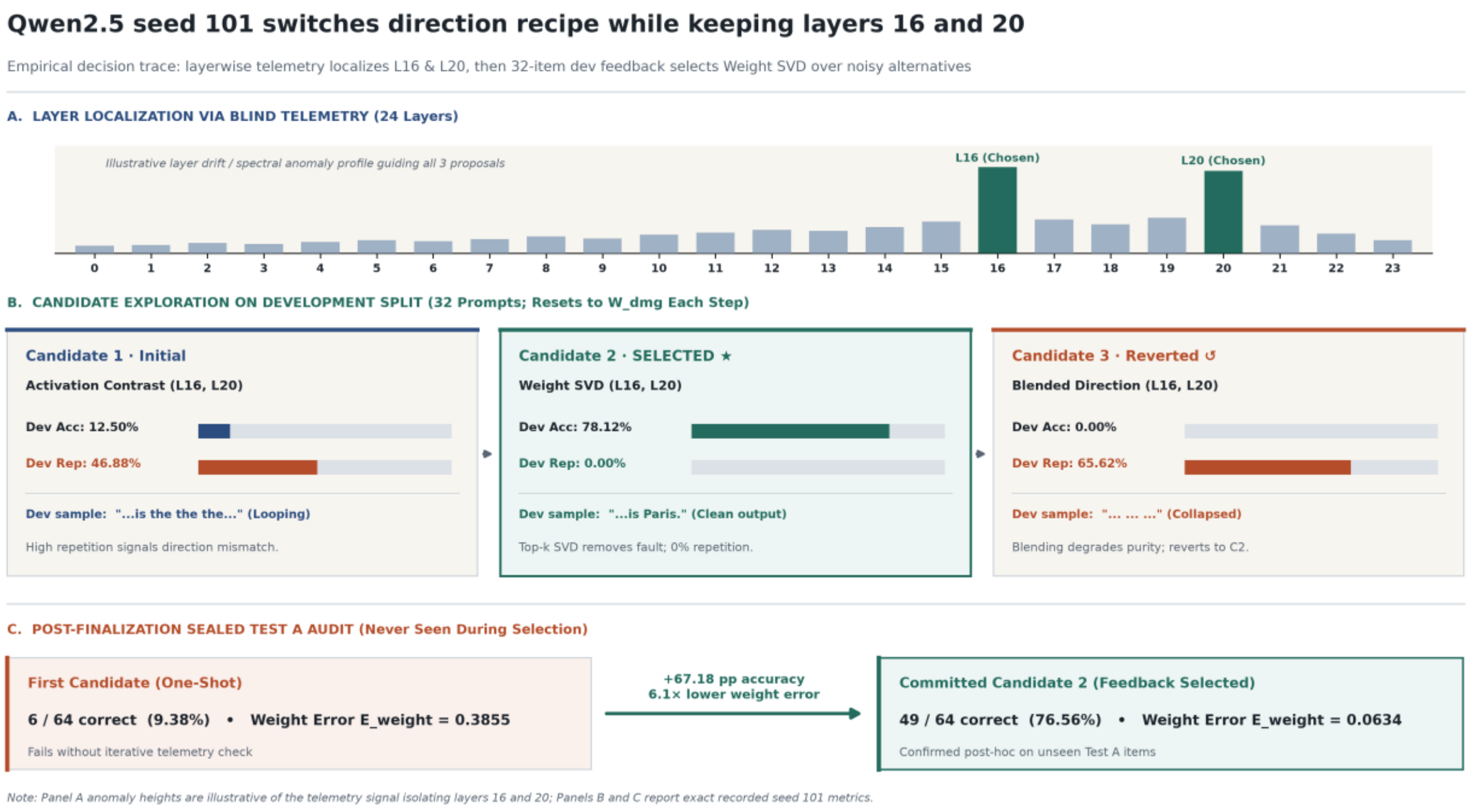

*Figure 7. Recorded Qwen feedback sequence. All three candidates used layers 16 and 20. Development scores select the spectral recipe; held-out scores are measured afterward. This shows how a controller can revise a poor initial direction choice without claiming a new projection operator.*

## 6.7 Fault support rank and layer coverage

*Table 7. Five additional single-seed SmolLM2 live sessions. The support, rank, and layer count varied separately. Each row is a diagnostic case rather than a multi-seed robustness estimate.*

| Fault | First A | Feedback A | Bayesian A | Feedback B | Weight error (Live / Bayes) | Layers |
|---|---|---|---|---|---|---|
| Random direction | 48/64 (75.0%) | 48/64 (75.0%) | 75.0% | 76.56 | 0.0671 / 0.0672 | 18,22 |
| Support 32 | 49/64 (76.6%) | 47/64 (73.4%) | 73.4% | 75.00 | 0.0644 / 0.0677 | 18,22 |
| Support 256 | 10/64 (15.6%) | 47/64 (73.4%) | 0.0% | 76.56 | 0.0813 / 0.9539 | 18,22 |

| Fault | First A | Feedback A | Bayesian A | Feedback B | Weight error (Live / Bayes) | Layers |
|---|---|---|---|---|---|---|
| Rank four | 49/64 (76.6%) | 47/64 (73.4%) | 56.2% | 75.00 | 0.1328 / 0.9048 | 18,22 |
| Four layers | 12/64 (18.8%) | 49/64 (76.6%) | 29.7% | 73.44 | 0.2637 / 0.7434 | 14,18,20,22 |

The additional live sessions test situations in which a fixed 64-column, two-layer edit may be insufficient. With 256 corrupted columns, the first SmolLM2 candidate scores 10/64; adaptive support raises the final score to 47/64. With four affected layers, the first candidate edits two layers and scores 12/64. The selected spectral edit covers layers 14, 18, 20, and 22 and reaches 49/64. Its weight error remains at 0.264, showing useful behavioral recovery without exact geometric restoration.

However, feedback is not uniformly helpful. The random-direction session retained 48/64, and the support-32 and rank-four sessions changed from 49/64 to 47/64 despite development-based selection. Five extra sessions on one seed are diagnostic demonstrations of capacity and selection effects, not a broad population robustness estimates.

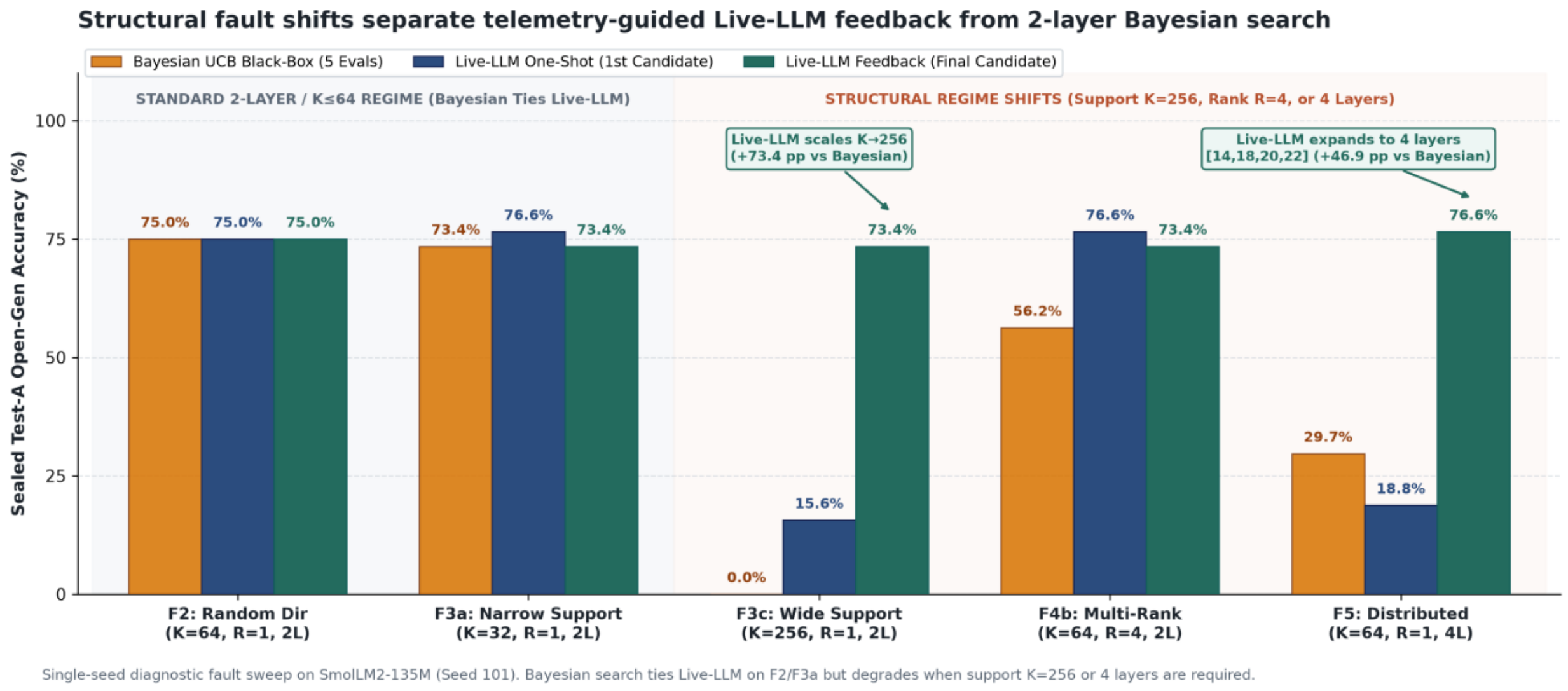

*Figure 8. Five non-core SmolLM2 fault regimes comparing matched-budget Bayesian/UCB black-box search (five evaluations) against Live-LLM one-shot and feedback candidates. While the Bayesian search matches Live-LLM on two-layer K≤64 faults (F2, F3a), telemetry-guided Live-LLM feedback outperforms the black-box search under support-width (F3c, K=256), rank (F4b, R=4), and four-layer (F5) shifts.*

*Table 8. Scripted fault sweep on one seed. The fixed projection edits two layers, whereas the scripted search may select four. Scores are percent correct. These are algorithmic controls and are separate from the five additional Live LLM sessions.*

| Target | Fault | K | Rank | Damaged A | Fixed A | Bayesian A | Scripted A |
|---|---|---|---|---|---|---|---|
| BERT | Aligned rank one | 64 | 1 | 65.62 | 62.50 | 65.62 | 68.75 |
| BERT | Random direction | 64 | 1 | 51.56 | 62.50 | 60.94 | 60.94 |
| BERT | Support 32 | 32 | 1 | 64.06 | 62.50 | 65.62 | 65.62 |
| BERT | Support 128 | 128 | 1 | 51.56 | 67.19 | 62.50 | 60.94 |
| BERT | Support 256 | 256 | 1 | 0.00 | 9.38 | 46.88 | 60.94 |
| BERT | Rank two | 64 | 2 | 67.19 | 65.62 | 60.94 | 56.25 |

| Target | Fault | K | Rank | Damaged A | Fixed A | Bayesian A | Scripted A |
|---|---|---|---|---|---|---|---|
| BERT | Rank four | 64 | 4 | 59.38 | 64.06 | 68.75 | 57.81 |
| BERT | Four layers | 64 | 1 | 53.12 | 40.62 | 71.88 | 67.19 |
| PubMed BERT | Aligned rank one | 64 | 1 | 76.56 | 70.31 | 75.00 | 73.44 |
| PubMed BERT | Random direction | 64 | 1 | 62.50 | 68.75 | 68.75 | 71.88 |
| PubMed BERT | Support 32 | 32 | 1 | 71.88 | 68.75 | 68.75 | 68.75 |
| PubMed BERT | Support 128 | 128 | 1 | 68.75 | 70.31 | 73.44 | 70.31 |
| PubMed BERT | Support 256 | 256 | 1 | 59.38 | 18.75 | 62.50 | 71.88 |
| PubMed BERT | Rank two | 64 | 2 | 75.00 | 71.88 | 73.44 | 71.88 |
| PubMed BERT | Rank four | 64 | 4 | 68.75 | 67.19 | 68.75 | 67.19 |
| PubMed BERT | Four layers | 64 | 1 | 7.81 | 70.31 | 64.06 | 71.88 |
| SmolLM2 | Aligned rank one | 64 | 1 | 6.25 | 75.00 | 75.00 | 75.00 |
| SmolLM2 | Random direction | 64 | 1 | 48.44 | 75.00 | 75.00 | 75.00 |
| SmolLM2 | Support 32 | 32 | 1 | 1.56 | 76.56 | 73.44 | 73.44 |
| SmolLM2 | Support 128 | 128 | 1 | 0.00 | 50.00 | 31.25 | 73.44 |
| SmolLM2 | Support 256 | 256 | 1 | 0.00 | 15.62 | 0.00 | 73.44 |
| SmolLM2 | Rank two | 64 | 2 | 7.81 | 60.94 | 60.94 | 71.88 |
| SmolLM2 | Rank four | 64 | 4 | 26.56 | 76.56 | 56.25 | 76.56 |
| SmolLM2 | Four layers | 64 | 1 | 3.12 | 4.69 | 29.69 | 71.88 |
| Pythia | Aligned rank one | 64 | 1 | 0.00 | 67.19 | 0.00 | 67.19 |
| Pythia | Random direction | 64 | 1 | 0.00 | 67.19 | 12.50 | 67.19 |
| Pythia | Support 32 | 32 | 1 | 0.00 | 6.25 | 68.75 | 68.75 |
| Pythia | Support 128 | 128 | 1 | 0.00 | 0.00 | 0.00 | 18.75 |
| Pythia | Support 256 | 256 | 1 | 0.00 | 0.00 | 3.12 | 0.00 |
| Pythia | Rank two | 64 | 2 | 0.00 | 4.69 | 0.00 | 17.19 |
| Pythia | Rank four | 64 | 4 | 0.00 | 0.00 | 0.00 | 3.12 |
| Pythia | Four layers | 64 | 1 | 0.00 | 0.00 | 0.00 | 0.00 |
| Qwen2.5 | Aligned rank one | 64 | 1 | 0.00 | 9.38 | 76.56 | 73.44 |
| Qwen2.5 | Random direction | 64 | 1 | 25.00 | 76.56 | 76.56 | 76.56 |
| Qwen2.5 | Support 32 | 32 | 1 | 0.00 | 78.12 | 78.12 | 78.12 |
| Qwen2.5 | Support 128 | 128 | 1 | 0.00 | 0.00 | 23.44 | 75.00 |
| Qwen2.5 | Support 256 | 256 | 1 | 0.00 | 0.00 | 28.12 | 78.12 |
| Qwen2.5 | Rank two | 64 | 2 | 0.00 | 64.06 | 84.38 | 75.00 |

| Target | Fault | K | Rank | Damaged A | Fixed A | Bayesian A | Scripted A |
|---|---|---|---|---|---|---|---|
| Qwen2.5 | Rank four | 64 | 4 | 4.69 | 42.19 | 60.94 | 67.19 |
| Qwen2.5 | Four layers | 64 | 1 | 0.00 | 0.00 | 59.38 | 20.31 |

The full non-LLM sweep adds important context. Wider support and distributed damage can sharply reduce fixed projection performance, and the scripted search can recover some cases by changing the support or layer scope. However, it is not uniformly successful, especially for Pythia's wider faults. Its ability to choose four layers, whereas the fixed method chooses two, is an explicit capacity advantage. Adaptive support and oracle rank diagnostics should likewise be labeled according to any extra information they receive.

## 6.8 Training duration changes the comparison

*Table 9. Saved three-seed horizon aggregates across all five targets (25, 50, and 100 AdamW steps). The values show the percent mean and sample SD.*

| Target | Steps | LoRA | Position repair then LoRA | Live LLM repair then LoRA |
|---|---|---|---|---|
| BERT | 25 | 97.40 ± 2.39 | 96.88 ± 0.00 | 96.36 ± 0.91 |
| BERT | 50 | 96.88 ± 1.57 | 96.88 ± 1.57 | 96.36 ± 2.39 |
| BERT | 100 | 95.83 ± 2.39 | 96.88 ± 0.00 | 96.88 ± 0.00 |
| PubMed BERT | 25 | 83.85 ± 9.15 | 84.37 ± 4.13 | 84.37 ± 4.13 |
| PubMed BERT | 50 | 88.02 ± 1.80 | 85.94 ± 0.00 | 85.42 ± 0.90 |
| PubMed BERT | 100 | 86.98 ± 2.38 | 86.46 ± 0.90 | 86.46 ± 0.90 |
| SmolLM2 | 25 | 86.98 ± 5.92 | 92.71 ± 4.78 | 95.32 ± 2.71 |
| SmolLM2 | 50 | 96.88 ± 3.13 | 96.88 ± 1.57 | 96.36 ± 0.91 |
| SmolLM2 | 100 | 96.35 ± 3.25 | 96.88 ± 0.00 | 96.36 ± 0.91 |
| Pythia | 25 | 0.00 ± 0.00 | 55.73 ± 38.82 | 71.35 ± 12.14 |
| Pythia | 50 | 0.00 ± 0.00 | 83.33 ± 4.51 | 76.04 ± 13.29 |
| Pythia | 100 | 19.79 ± 34.28 | 84.38 ± 1.57 | 80.21 ± 6.31 |
| Qwen2.5 | 25 | 89.58 ± 6.31 | 100.00 ± 0.00 | 100.00 ± 0.00 |
| Qwen2.5 | 50 | 95.84 ± 3.25 | 99.48 ± 0.90 | 98.96 ± 1.80 |
| Qwen2.5 | 100 | 98.44 ± 0.00 | 99.48 ± 0.90 | 99.48 ± 0.90 |

Across all five targets, longer training horizons clarify when prior structural repair is essential and when the standalone LoRA eventually catches up. On Pythia-70M, sequential repair-plus-LoRA rises from 55.73% at 25 steps to 83.33% at 50 and 84.38% at 100 steps (with Live-LLM repair-plus-LoRA rising from 71.35% to 80.21%), whereas standalone LoRA remains collapsed at 0.00% through 50 steps and reaches only 19.79% at 100 steps. On Qwen2.5-0.5B, both Fixed Projection + LoRA and Live-LLM Repair + LoRA reach 100.00% immediately at 25 steps (and maintain 99.48% at 100 steps), whereas standalone LoRA starts at 89.58% at 25 steps and gradually climbs to 95.83% at 50 steps and 98.44% at 100 steps.

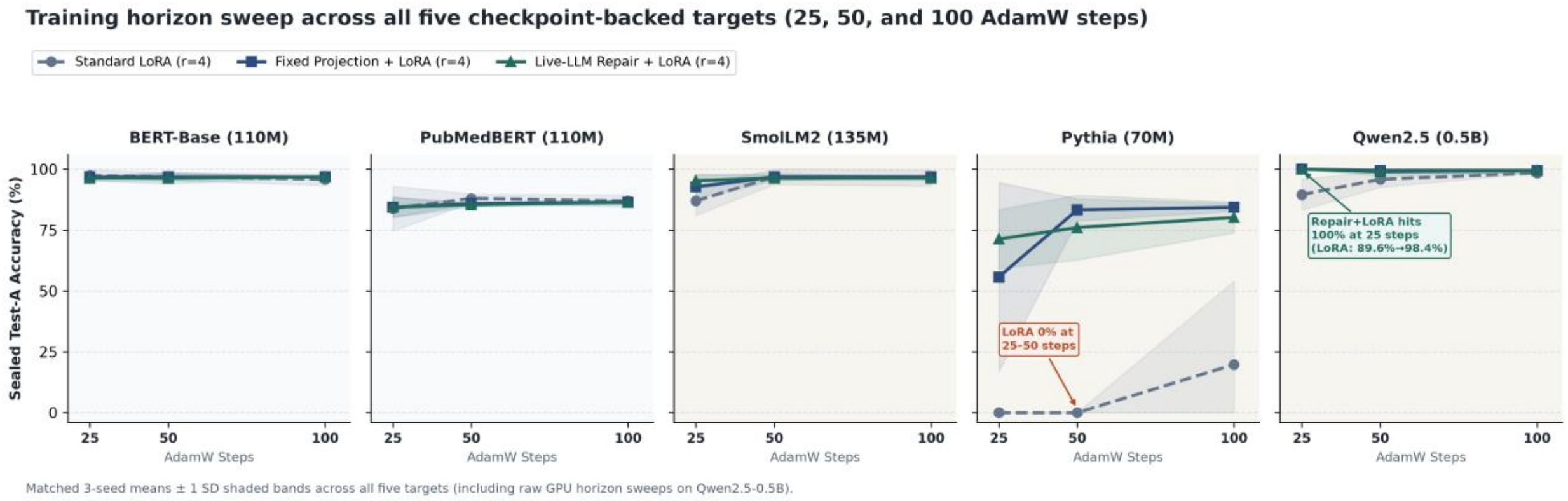

*Figure 9. Training-duration horizon sweep (25, 50, and 100 AdamW steps) across all five checkpoint-backed targets (including Qwen2.5-0.5B). Clearing the fault prior to LoRA prevented short-horizon collapse on Pythia-70M and reached 100.0% at 25 steps on Qwen2.5-0.5B.*

# 7 Exploratory recovery without implanted damage

This study began with unmodified causal checkpoints. There are 32 development prompts, repartitioned in three ways into 16 diagnostic and 16 validation prompts, and 24 base test questions under four framing prefixes. The resulting 96 prompts measure framing sensitivity; they are not 96 independent questions. This is a non-LLM projection study rather than an additional live-controlled experiment. Its localizer collects failure and passing counts but returns a position-contrast plan; it does not implement a separate hidden-state difference between the two groups.

The projection strength is selected from {0,0.25,0.5,0.75,1} on validation completion accuracy minus half the repetition rate, with retention loss as a tie-breaker. Including zero permits abstention. The test scores do not choose the strength. If the diagnostic set has no mixture of failing and passing trajectories, localization falls back to a prediction-versus-context contrast. SmolLM2 has zero flagged diagnostic failures in all three partitions; thus, fallback is used throughout.

*Table 10. Natural completion without implanted faults. The 96 variants represent four frames of 24 base questions. The penalty values are fixed controls and not validation-matched searches.*

| Model | Condition | Correct / 96 | Flags / 96 | PPL |
|---|---|---|---|---|
| Pythia | Original | 10 | 5 | 172.80 |
| Pythia | Penalty 1.35 | 5 | 9 | 172.80 |
| Pythia | Projection 0.5 | 10 | 4 | 155.62 |
| SmolLM2 | Original | 29 | 4 | 61.16 |
| SmolLM2 | Penalty 1.35 | 22 | 5 | 61.16 |
| SmolLM2 | Projection 0.5 | 28 | 5 | 70.80 |
| Qwen2.5 | Original | 45 | 0 | 20.75 |
| Qwen2.5 | Penalty 1.35 | 33 | 2 | 20.75 |
| Qwen2.5 | Projection 0.5 | 45 | 0 | 20.97 |

*Table 11. Natural edits selected on separate validation prompts, including alpha zero. Three development resamples reused the same test questions. Gains and losses were paired variant counts relative to the original outputs.*

| Model | Split seed | Alpha | Correct / 96 | Flags / 96 | Gains / losses |
|---|---|---|---|---|---|
| Pythia | 101 | 0.75 | 8 | 7 | 1 / 3 |
| Pythia | 202 | 1.0 | 5 | 14 | 1 / 6 |
| Pythia | 303 | 1.0 | 5 | 12 | 1 / 6 |
| SmolLM2 | 101 | 0.5 | 28 | 5 | 2 / 3 |
| SmolLM2 | 202 | 0.5 | 28 | 5 | 2 / 3 |
| SmolLM2 | 303 | 0.0 | 29 | 4 | 0 / 0 |
| Qwen2.5 | 101 | 1.0 | 46 | 1 | 1 / 0 |
| Qwen2.5 | 202 | 0.5 | 46 | 0 | 1 / 0 |
| Qwen2.5 | 303 | 1.0 | 47 | 1 | 2 / 0 |

The tuned Qwen edits added one, one, and two correct variants with no lost correct variants, changing the mean from 45/96 to 46.33/96. These are small changes on repeated framings and require grouping by base question for inference. SmolLM2 changes from 29/96 to 28, 28, and 29, abstaining in one partition, respectively. Pythia changed from 10/96 to 8, 5, and 5. A fixed half-strength Pythia edit preserves the original ten correct variants and reduces one repetition flag, but validation-selected stronger edits perform worse on the test.

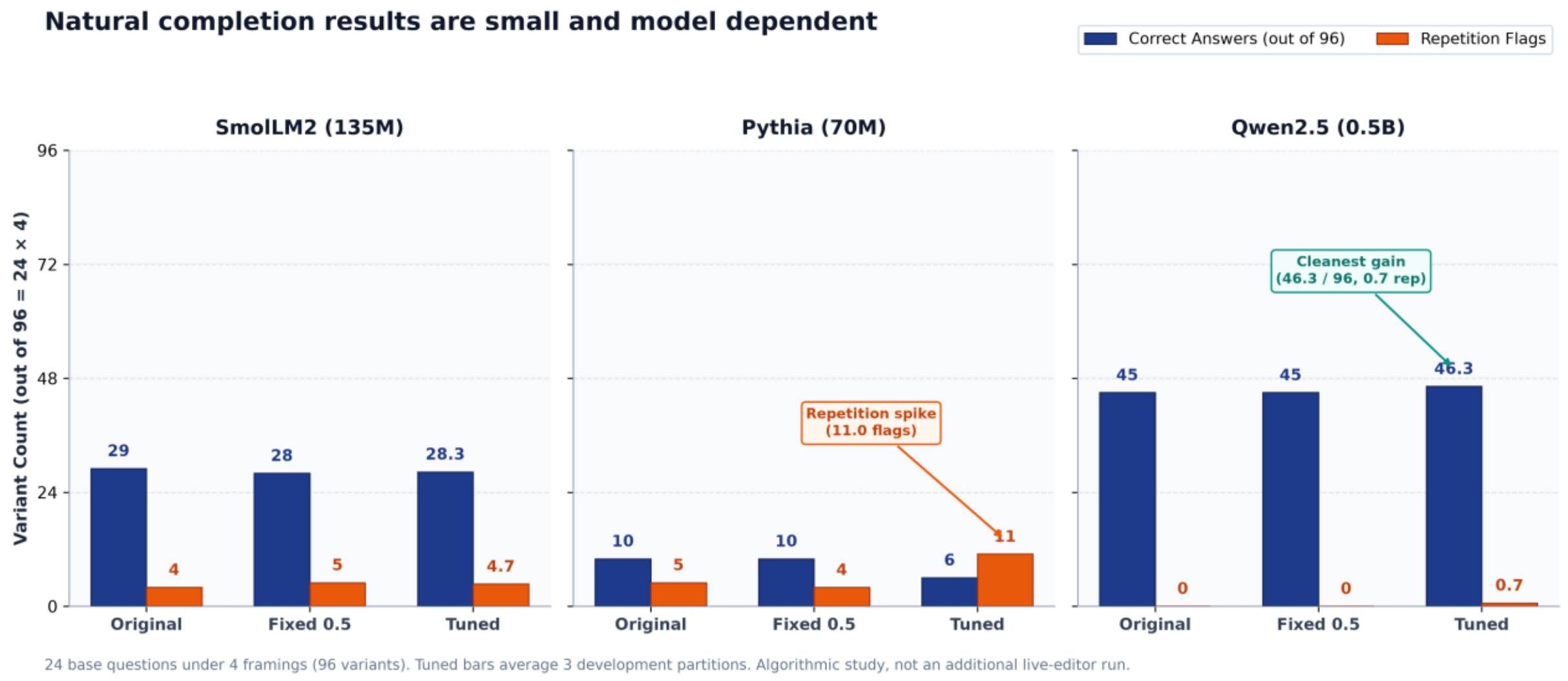

*Figure 10. Natural-completion counts on 24 base questions under four framings. Tuned values average development partitions on the same 96 variants. Small Qwen gains coexist with losses on other targets; therefore, this study remains exploratory.*

The artifact stores per-item correctness and flag vectors, with sample strings for selected conditions rather than complete token histories for every output. Exact first-word scoring can reject reasonable alternative phrasings or numeric forms. All outputs, normalized acceptable answers, and a larger set of independent questions are needed before making a broad natural-recovery claim.

# 8 Repair of compound attention and MLP faults

The core experiments vary the parameters of a single MLP projection family. The compound experiment adds three operators with snapshot and readback verification. One edits an attention output head by projection or ablation, another rescales the MLP output matrices, and the third projects selected MLP columns. The experiment tests whether a controller can compose operators that address distinct types of damage.

We evaluated this operator family under a controlled 3-component compound fault (F6_Compound_AttnHead_Scale_MLP: a rank-2 spike in one attention head, full-rank 2.45 × 3.20 × gain drift across two MLP layers, and a sparse K=64 rank-1 MLP column spike) across five Transformer checkpoints—two masked encoders (BERT-Base-Uncased, PubMedBERT-Base) and three causal decoders (Pythia-70M-Deduped, SmolLM2-135M-Instruct, Qwen2.5-0.5B-Instruct)—across three paired fault seeds (101, 202, 303; 15 model-seed cases). Because sparse MLP column projection cannot fix the attention-head spike or full-rank gain drift, both MLP-Only WSP64 (Arm 04) and MLP-Only similarity-based UCB search over 5 dev evaluations (Arm 28) fail: 5-checkpoint descriptive mean Test-A accuracy remains at 25.31% and 25.31% (vs. 24.89% damaged and 71.87% original clean), with combined weight error E_weight,total = 0.9355 and 0.9626.

The LLM-proposed programs identified the injected locations in all 15 model-seed cases and achieved a mean Test-A accuracy of 71.25 % %with a relative weight error of 0.0508. The scripted selector achieved the same values when evaluating the matched set of three candidates. The first LLM proposal achieved 70.31%. Test-B gains varied across targets, with Pythia recovering from 1.04% to 12.50%, respectively. The compound-fault comparison supports the contribution of the operator family and development selection under these conditions.

*Table 12. Compound-fault test-A accuracy and relative weight error across three paired seeds per checkpoint. The parentheses show the relative weight error. The final column shows the descriptive mean across the five checkpoint-specific scores.*

| Editing Arm / Method | BERT-Base (Enc) | PubMedBERT (Enc) | Pythia-70M (Dec) | SmolLM2-135M (Dec) | Qwen2.5-0.5B (Dec) | 5-Ckpt Mean (Ew) |
|---|---|---|---|---|---|---|
| **Original Clean (00)** | 65.6±0.0% (0.000) | 71.9±0.0% (0.000) | 70.3±0.0% (0.000) | 75.0±0.0% (0.000) | 76.6±0.0% (0.000) | 71.9% (0.000) |
| **Compound Damaged (01)** | 4.2±3.2% (1.000) | 5.2±4.5% (1.000) | 1.0±1.8% (1.000) | 67.7±4.8% (1.000) | 46.4±5.0% (1.000) | 24.9% (1.000) |
| **MLP-Only WSP64 (04)** | 5.2±6.5% (0.924) | 0.0±0.0% (0.932) | 1.6±2.7% (0.919) | 71.9±4.1% (0.930) | 47.9±3.2% (0.973) | 25.3% (0.936) |
| **MLP-Only Sim-UCB 5-Eval (28)** | 8.8±10.0% (0.924) | 0.0±0.0% (0.977) | 1.6±2.7% (0.999) | 69.3±0.9% (0.930) | 46.9±5.6% (0.982) | 25.3% (0.963) |
| **Std LoRA MLP r=4, 25st (10)** | 60.9±4.1% (1.000) | 52.6±2.4% (1.000) | 20.3±10.2% (1.000) | 88.5±5.5% (1.000) | 94.8±3.6% (1.000) | 63.4% (1.000) |
| **Std LoRA Attn+MLP 25st (10b)** | 54.7±5.6% (1.000) | 53.1±1.6% (1.000) | 22.9±8.0% (1.000) | 91.7±2.4% (1.000) | 93.8±7.2% (1.000) | 63.2% (1.000) |
| **Matched Scripted 3-Cand (29)** | 65.1±0.9% (0.015) | 72.4±0.9% (0.031) | 63.5±0.9% (0.117) | 74.5±0.9% (0.043) | 80.7±3.3% (0.047) | 71.2% (0.051) |
| **LLM-Proposed 1-Shot (30)** | 65.1±0.9% (0.015) | 71.4±0.9% (0.017) | 63.5±0.9% (0.110) | 74.5±0.9% (0.027) | 77.1±0.9% (0.024) | 70.3% (0.039) |
| **LLM-Proposed Dev-Sel (31)** | 65.1±0.9% (0.015) | 72.4±0.9% (0.031) | 63.5±0.9% (0.117) | 74.5±0.9% (0.043) | 80.7±3.3% (0.047) | 71.2% (0.051) |

| Editing Arm / Method | BERT-Base (Enc) | PubMedBERT (Enc) | Pythia-70M (Dec) | SmolLM2-135M (Dec) | Qwen2.5-0.5B (Dec) | 5-Ckpt Mean (Ew) |
|---|---|---|---|---|---|---|
| **Matched Scripted + LoRA 25st** | 81.8±3.3% (0.017) | 78.6±2.4% (0.032) | 74.0±0.9% (0.118) | 97.9±0.9% (0.044) | 97.4±2.4% (0.056) | 85.9% (0.053) |
| **LLM-Proposed + LoRA 25st (32)** | 81.8±3.3% (0.017) | 78.6±2.4% (0.032) | 74.0±0.9% (0.118) | 97.9±0.9% (0.044) | 97.4±2.4% (0.056) | 85.9% (0.053) |

The eight-subset ablation in Table 13 applies exact subsets of the selected programs. The full program reached a mean Test-A accuracy of 71.25 %, compared with 64.38% for the best two-operator subset and 24.89% in the damaged state. It also lowers mean relative weight error to 0.0508. The full set need not maximize each model's task score. On Qwen2.5, attention repair plus scale correction scores 81.77% versus 80.73% for all three operators, while adding MLP projection reduces weight error from 0.2347 to 0.0467.

With balanced 32-example training and rank-four LoRA, LLM-selected repair followed by adaptation reaches 85.94% at 25 steps and 94.48% at 500 steps. LoRA on the damaged attention and MLP weights reached 63.23% and 82.92% at the same horizon. The repair-plus-LoRA arm trains adapters on the MLP modules, while the attention-plus-MLP baseline uses a broader adapter scope. These comparisons apply to the compound fault and task in this experiment.

*Table 13. Exact operator-subset ablations and balanced 32-example LoRA horizon sweep at 25, 100, 200, and 500 AdamW steps.*

| Editing Arm / Method | BERT-Base (Enc) | PubMedBERT (Enc) | Pythia-70M (Dec) | SmolLM2-135M (Dec) | Qwen2.5-0.5B (Dec) | 5-Ckpt Mean (Ew) |
|---|---|---|---|---|---|---|
| **Subset: None (Damaged)** | 4.2±3.2% (1.000) | 5.2±4.5% (1.000) | 1.0±1.8% (1.000) | 67.7±4.8% (1.000) | 46.4±5.0% (1.000) | 24.9% (1.000) |
| **Subset: Op3 Only (MLP Exact)** | 11.5±13.3% (0.927) | 4.7±2.7% (0.936) | 1.6±2.7% (0.923) | 69.3±3.9% (0.930) | 49.5±1.8% (0.973) | 27.3% (0.938) |
| **Subset: Op1 Only (Attn Head)** | 32.3±7.9% (0.940) | 15.6±8.7% (0.907) | 2.1±0.9% (0.946) | 61.5±0.9% (0.923) | 49.0±3.2% (0.957) | 32.1% (0.935) |
| **Subset: Op1 + Op3** | 41.1±2.4% (0.862) | 21.4±0.9% (0.836) | 4.7±0.0% (0.864) | 64.6±2.4% (0.848) | 53.6±3.9% (0.929) | 37.1% (0.868) |
| **Subset: Op2 Only (Scale)** | 37.5±23.6% (0.506) | 38.0±19.1% (0.550) | 6.8±11.7% (0.518) | 68.2±3.3% (0.533) | 80.7±2.4% (0.373) | 46.2% (0.496) |
| **Subset: Op2 + Op3** | 41.7±29.6% (0.340) | 47.4±10.2% (0.421) | 2.1±2.4% (0.346) | 75.0±4.1% (0.387) | 80.7±2.4% (0.294) | 49.4% (0.357) |
| **Subset: Op1 + Op2** | 62.5±5.6% (0.375) | 61.5±7.4% (0.354) | 43.8±13.9% (0.403) | 72.4±1.8% (0.369) | 81.8±4.5% (0.235) | 64.4% (0.347) |
| **Subset: All 3 Ops (Selected)** | 65.1±0.9% (0.015) | 72.4±0.9% (0.031) | 63.5±0.9% (0.117) | 74.5±0.9% (0.043) | 80.7±3.3% (0.047) | 71.2% (0.051) |
| **Std LoRA Attn+MLP (25st)** | 54.7±5.6% (1.000) | 53.1±1.6% (1.000) | 22.9±8.0% (1.000) | 91.7±2.4% (1.000) | 93.8±7.2% (1.000) | 63.2% (1.000) |
| **Std LoRA Attn+MLP (100st)** | 70.3±9.4% (1.000) | 57.8±4.1% (1.000) | 38.5±4.5% (1.000) | 93.8±4.1% (1.000) | 99.5±0.9% (1.001) | 72.0% (1.000) |
| **Std LoRA Attn+MLP (200st)** | 82.3±9.5% (1.000) | 65.6±6.2% (1.000) | 44.3±13.3% (1.000) | 93.2±2.4% (1.000) | 99.5±0.9% (1.001) | 77.0% (1.000) |
| **Std LoRA Attn+MLP (500st)** | 92.2±4.1% (1.000) | 86.5±3.9% (1.001) | 44.3±11.8% (1.000) | 92.7±2.4% (1.000) | 99.0±1.8% (1.001) | 82.9% (1.001) |
| **Matched Scripted + LoRA (25st)** | 81.8±3.3% (0.017) | 78.6±2.4% (0.032) | 74.0±0.9% (0.118) | 97.9±0.9% (0.044) | 97.4±2.4% (0.056) | 85.9% (0.053) |

| Editing Arm / Method | BERT-Base (Enc) | PubMedBERT (Enc) | Pythia-70M (Dec) | SmolLM2-135M (Dec) | Qwen2.5-0.5B (Dec) | 5-Ckpt Mean (Ew) |
|---|---|---|---|---|---|---|
| **Matched Scripted + LoRA (500st)** | 95.8±1.8% (0.031) | 91.1±2.4% (0.045) | 89.1±2.7% (0.119) | 97.9±0.9% (0.045) | 99.0±0.9% (0.061) | 94.6% (0.060) |
| **LLM-Proposed + LoRA (25st)** | 81.8±3.3% (0.017) | 78.6±2.4% (0.032) | 74.0±0.9% (0.118) | 97.9±0.9% (0.044) | 97.4±2.4% (0.056) | 85.9% (0.053) |
| **LLM-Proposed + LoRA (100st)** | 95.8±2.4% (0.023) | 92.2±1.6% (0.036) | 89.6±0.9% (0.119) | 98.4±0.0% (0.045) | 99.5±0.9% (0.061) | 95.1% (0.057) |
| **LLM-Proposed + LoRA (200st)** | 94.3±0.9% (0.028) | 91.1±3.9% (0.041) | 89.6±3.6% (0.119) | 98.4±0.0% (0.045) | 99.0±0.9% (0.061) | 94.5% (0.059) |
| **LLM-Proposed + LoRA (500st)** | 95.8±1.8% (0.032) | 90.6±2.7% (0.046) | 89.1±2.7% (0.119) | 97.9±0.9% (0.045) | 99.0±0.9% (0.061) | 94.5% (0.061) |

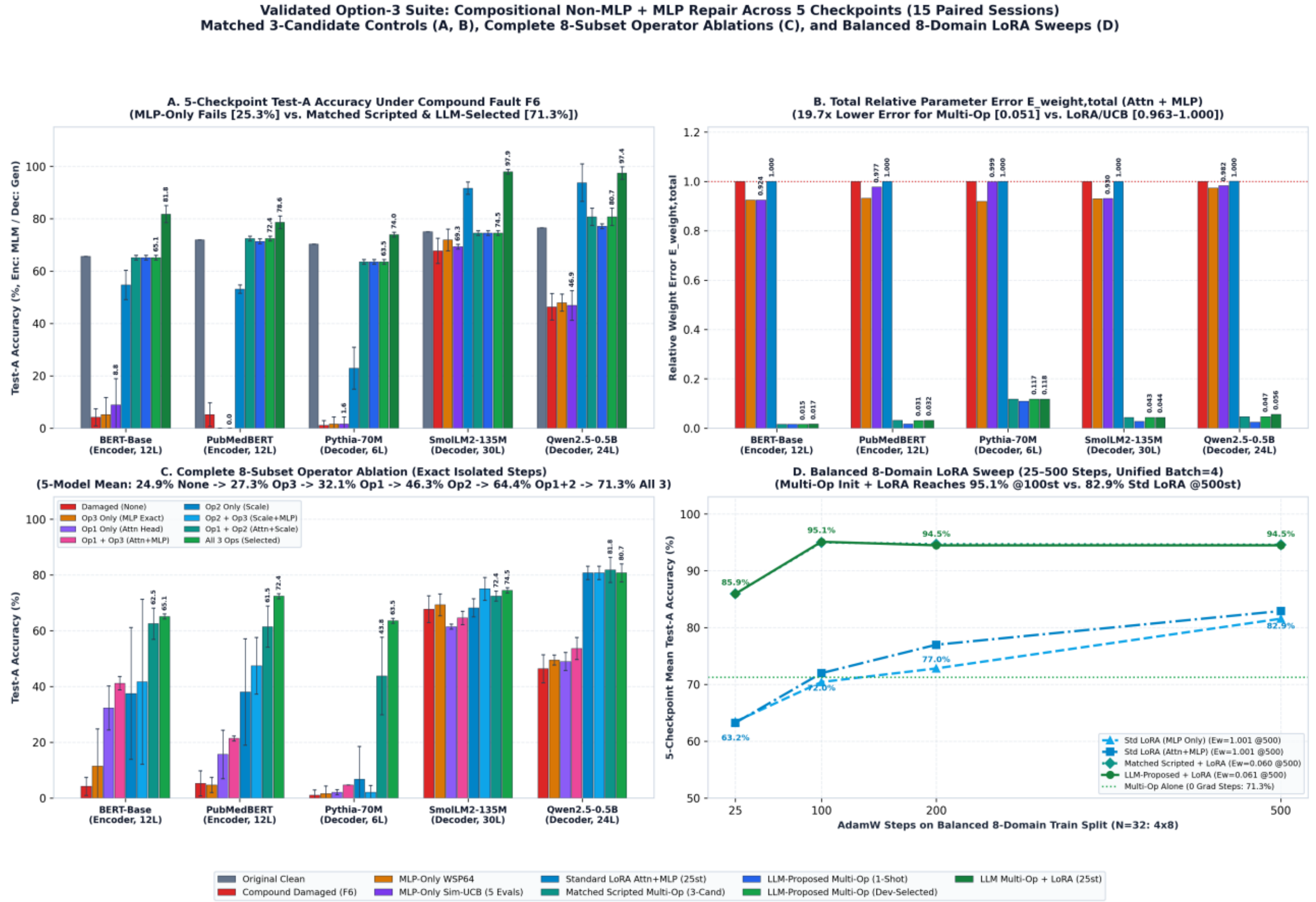

*Figure 11. Validated Option-3 Multi-Operator Benchmark across five transformer checkpoints (three paired seeds = 15 cases). (A) Test-A accuracy showing MLP-only failure (25.3%) vs. matched scripted and LLM-proposed recovery (71.3%). (B) Combined attention + MLP relative parameter error E_weight,total. (C) Complete 8-subset operator ablation using exact isolated steps. (D) Unified balanced 8-domain LoRA horizon sweep (25–500 steps, batch=4).*

# 9 Repair after training and quantization failures

We evaluate failures induced by four procedures without directly adding a sparse matrix spike. The procedures were off-domain fine-tuning, high-learning-rate fine-tuning, supervised fine-tuning with flipped labels, and simulated low-bit quantization. Each was evaluated on four checkpoints and three seeds, for 48 model–fault–seed cases. A separate Pythia experiment recorded eight sequential editor sessions, with two seeds per failure procedure and up to five development evaluations per session. The quantization configuration differs for Qwen2.5, which uses per-row four-bit quantization. The other targets use per-tensor three-bit quantization.

Across the three training-induced failures, the mean Test-A accuracy after scripted spectral or scale-based repair rounds was within 0.6 percentage points of the failed-state mean. The 25-step LoRA reached 74.6%, 76.8%, and 71.9% on those conditions, respectively. In the two-seed Pythia quantization comparison, sequential LLM selection reached 48.44% compared with the reported 39.06% scripted control. LLM-guided repair followed by LoRA reached 72.66% compared with 55.47% for standalone LoRA in that comparison. The four-model matrix and sequential experiment used different seed counts and were reported separately.

*Table 14. Test-A scores for four induced failures across four targets and three seeds per condition. The four-model mean combines checkpoint-specific masked-token and generation metrics. The separate two-seed Pythia comparison is presented in the text and Figure 12.*

| Failure and intervention | BERT-Base (Clean 65.6%) | Pythia-70M (Clean 70.3%) | SmolLM2-135M (Clean 75.0%) | Qwen2.5-0.5B (Clean 76.6%) | 4-Model Mean |
|---|---|---|---|---|---|
| 1. Off-Domain FT (Forgetting) \| Failed | 0.5% | 54.7% | 75.0% | 9.4% | 34.9% |
| 1. Off-Domain FT (Forgetting) \| Matched Scripted (5-Eval) | 0.0% | 56.2% | 76.6% | 9.4% | 35.5% |
| 1. Off-Domain FT (Forgetting) \| Std LoRA (25st) | 55.7% | 65.1% | 93.8% | 83.9% | 74.6% |
| 1. Off-Domain FT (Forgetting) \| Scripted + LoRA (25st) | 55.7% | 58.9% | 93.8% | 83.3% | 72.9% |
| 2. GradSpike FT (Collapse) \| Failed | 68.2% | 45.3% | 82.8% | 44.3% | 60.2% |
| 2. GradSpike FT (Collapse) \| Matched Scripted (5-Eval) | 68.2% | 47.4% | 83.3% | 44.3% | 60.8% |
| 2. GradSpike FT (Collapse) \| Std LoRA (25st) | 82.8% | 66.1% | 98.4% | 59.9% | 76.8% |
| 2. GradSpike FT (Collapse) \| Scripted + LoRA (25st) | 82.3% | 54.7% | 98.4% | 67.2% | 75.7% |
| 3. Poisoned SFT (Flipped Labels) \| Failed | 51.6% | 45.3% | 66.1% | 40.1% | 50.8% |
| 3. Poisoned SFT (Flipped Labels) \| Matched Scripted (5-Eval) | 52.1% | 46.9% | 65.6% | 38.5% | 50.8% |
| 3. Poisoned SFT (Flipped Labels) \| Std LoRA (25st) | 73.4% | 64.1% | 94.8% | 55.2% | 71.9% |
| 3. Poisoned SFT (Flipped Labels) \| Scripted + LoRA (25st) | 73.4% | 62.0% | 93.2% | 39.6% | 67.1% |
| 4. Low-Bit Quant (3b / Row-4b) \| Failed | 64.1% | 35.9% | 0.0% | 48.4% | 37.1% |
| 4. Low-Bit Quant (3b / Row-4b) \| Matched Scripted (5-Eval) | 65.6% | 39.1% | 0.0% | 48.4% | 38.3% |
| 4. Low-Bit Quant (3b / Row-4b) \| Std LoRA (25st) | 76.6% | 57.3% | 2.1% | 95.3% | 57.8% |
| 4. Low-Bit Quant (3b / Row-4b) \| Scripted + LoRA (25st) | 76.6% | 74.5% | 5.2% | 95.3% | 62.9% |

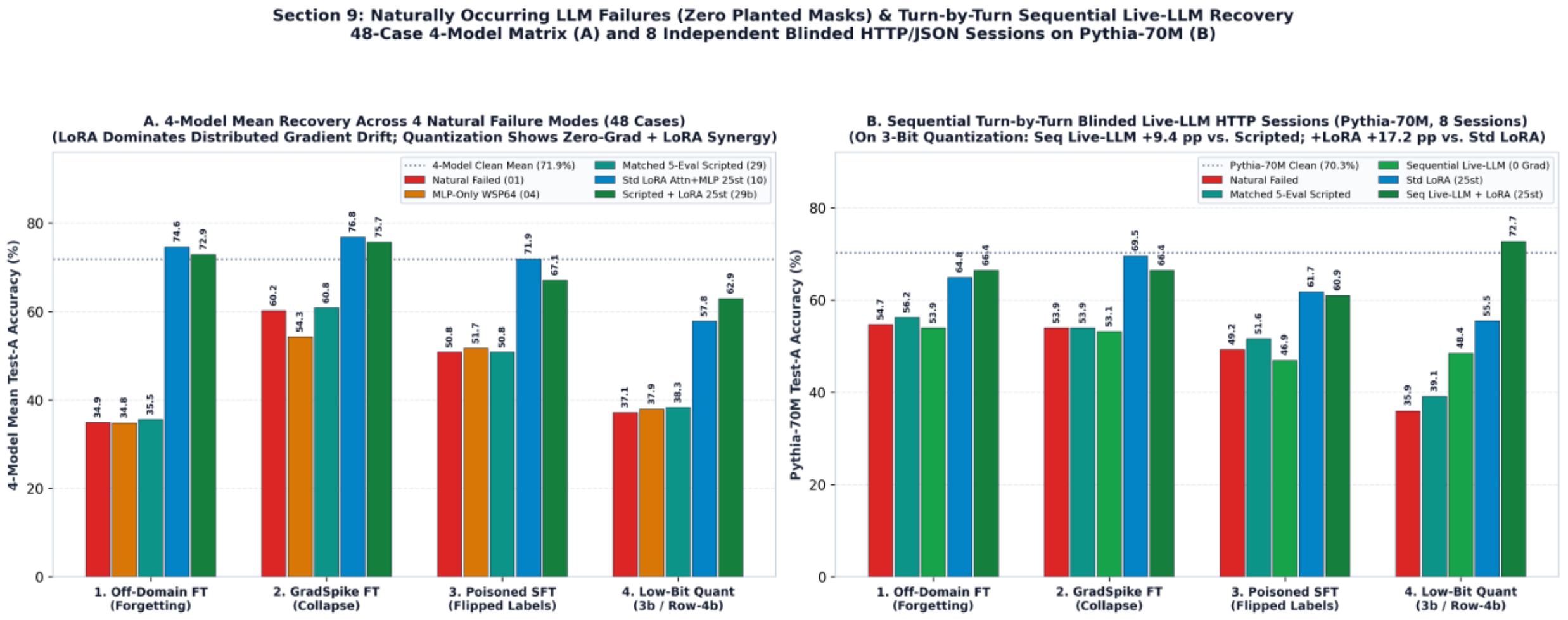

*Figure 12. Recovery after training-induced and quantization-induced failures. Panel A summarizes the four-model matrix. Panel B summarizes the eight Pythia sessions across four failure procedures with two seeds per procedure. The figure label "natural" denotes induced failures without a directly planted matrix spike. Section 7 separately studies unmodified checkpoints.*

### 9.1 Interpreting the quantization and adaptation results

Spectral clipping and scale corrections changed the repetition and completion behavior in the Pythia quantization runs. In the reported comparison, the repetition rate falls from 62.50% to 6.25% as the completion accuracy rises from 35.94% to 48.44%. This is consistent with the sensitivity to perturbations in weight geometry. Preservation of specific semantic directions and contribution of each operator would require separate measurements.

The scripted spectral and scale operators provide limited recovery for the training-induced conditions tested here. LoRA uses labeled task examples and yields higher completion scores in those comparisons. The experiments do not measure the rank of the training-induced weight changes or establish that they share a subspace with clean factual associations.

Taken together, the two extensions show that recovery depends on the failure procedure, available operators, and access to supervision. They provide no basis for treating any one repair family as a default across faults.

## 10 Discussion and limitations

The combined evidence supports direct weight editing as a useful route to recovery from controlled low-rank obstructions. Fixed non-LLM projections recover much of the causal core-fault behavior with a small residual geometric error. Live LLM editing adds an interactive way to choose among the same primitives, with recorded cases where development feedback changes an initially poor recipe or insufficient edit extent. The distinction is selection and composition under observations; the numerical projection is supplied by the environment.

The matched comparisons isolate two factors that can improve outcomes. Development feedback can change the selected recipe, as in Qwen's recorded sequence. A wider operator family can address multiple types of damage, as in the compound-fault experiment where the scripted and LLM-guided selectors tie. The findings apply to the stated benchmarks, control policies, and budgets. Reported weight recovery and short completion gains describe the properties of the edited model.

Several limitations define the scope. Faults are localized and deliberately low rank; therefore, the repair family is well matched to the construction. Task answers occupy eight adjective domains with informative lexical cues. The first-clause pairing blocks a specific lookup shortcut but does not remove every lexical strategy. Three or five fault seeds provide limited inferential precision, and the live first-versus-final comparison is retrospective within sessions. Statistical non-rejection is not evidence of equivalence.

The information boundary is an interface and an instruction protocol. The shared host state, descriptive initialization metadata, accessible audit action, and missing complete provider conversations prevent a stronger claim of independently verified secure blinding. Finalization is required before the audit; however, the source does not enforce a one-time immutable session lock, preventing further proposals after test feedback. There is no evidence that the recorded editors used test feedback for revision; the missing guard is a reproducibility and protocol limitation.

Method fidelity also matters. The rollout diagnostic direction currently remains a position contrast, live confidence-labeled application does not automatically apply scripted per-layer gates, and official-reference parity has not been demonstrated for all adaptation variants. The supervised training selector uses the first token rather than full-generation validation. Qwen's non-live comparison and adaptation rows lack the raw-run provenance required for inclusion. These limitations are stated in the method or result where they affect interpretation.

Future tests should preserve the exact editor revisions and complete conversations, enforce audit-only access after immutable commitment, save and verify committed weight deltas, and select adaptation checkpoints using the full generation objective. More independent faults, fewer projection-aligned perturbations, and broader completion tasks are needed to test the transfer. Larger preservation corpora and independent natural-failure questions are needed before claiming general capability recovery.

## 11 Conclusion

A tool-using language model can select effective repairs for controlled faults in transformer weights. Fixed and scripted methods perform comparably under several conditions, including the matched compound-fault comparison. Feedback helps select cases, and the benefit of repair before adaptation varies by target and failure procedure. Reporting retained behavior, weight distance, and held-out completions alongside each comparison makes those differences explicit.

# A Expanded seed summaries

The main text uses matched three-seed comparisons for every method. The following tables add two extra non-live fault seeds. Live rows continue to have three seeds. These summaries are supplementary coverage, not a replacement for the main comparison.

*Table 15. Expanded seed coverage for Test A. Non live methods use five fault seeds and live methods use three. These mixed seed summaries supplement the matched three-seed comparisons in the main text. The original checkpoint is the same reference in every fault trial.*

| Method | BERT | PubMed BERT | SmolLM2 | Pythia | Qwen2.5 |
|---|---|---|---|---|---|
| Original checkpoint n5 | 65.62 ± 0.00 | 71.88 ± 0.00 | 75.00 ± 0.00 | 70.31 ± 0.00 | 76.56 ± 0.00 |
| Damaged checkpoint n5 | 64.69 ± 2.84 | 71.25 ± 3.24 | 1.56 ± 2.71 | 0.00 ± 0.00 | 0.00 ± 0.00 |
| Repetition penalty n5 | 29.69 ± 5.63 | 10.00 ± 2.37 | 0.00 ± 0.00 | 0.00 ± 0.00 | 0.00 ± 0.00 |
| Oracle inverse n5 | 65.62 ± 0.00 | 71.88 ± 0.00 | 75.00 ± 0.00 | 70.31 ± 0.00 | 76.56 ± 0.00 |
| Position projection n5 | 62.81 ± 1.31 | 71.88 ± 1.57 | 74.06 ± 0.85 | 65.31 ± 2.57 | 62.50 ± 29.73 |
| Gated spectral projection n5 | 58.13 ± 1.31 | 72.19 ± 0.70 | 74.06 ± 0.85 | 65.31 ± 2.57 | 62.50 ± 29.73 |
| Dense position projection n5 | 62.81 ± 1.31 | 71.88 ± 1.11 | 74.06 ± 0.85 | 61.88 ± 3.76 | 75.62 ± 1.78 |
| Rollout diagnostic projection n5 | 62.81 ± 1.31 | 71.88 ± 1.57 | 74.06 ± 0.85 | 65.31 ± 2.57 | 62.50 ± 29.73 |
| Weight SVD projection n5 | 63.12 ± 1.78 | 71.25 ± 1.40 | 74.06 ± 0.85 | 0.00 ± 0.00 | 75.94 ± 1.40 |
| Context damping n5 | 62.19 ± 0.70 | 72.50 ± 1.78 | 73.13 ± 1.31 | 0.00 ± 0.00 | 59.38 ± 33.24 |
| LoRA n5 | 97.50 ± 1.78 | 83.44 ± 6.59 | 84.06 ± 5.89 | 0.00 ± 0.00 | 90.93 ± 5.11 |
| DoRA style n5 | 97.50 ± 1.78 | 85.94 ± 7.07 | 84.06 ± 5.68 | 0.00 ± 0.00 | 91.25 ± 5.91 |
| PiSSA style n5 | 93.44 ± 3.39 | 84.06 ± 6.09 | 84.37 ± 7.49 | 0.00 ± 0.00 | 96.25 ± 1.78 |
| LoReFT style n5 | 91.56 ± 6.01 | 89.37 ± 2.32 | 65.94 ± 23.63 | 0.00 ± 0.00 | 94.69 ± 1.78 |
| Support masked LoRA n5 | 82.81 ± 6.72 | 82.19 ± 7.13 | 80.63 ± 3.60 | 37.81 ± 8.37 | 91.87 ± 2.80 |
| Position repair then LoRA r3 n5 | 93.44 ± 3.39 | 84.37 ± 5.52 | 95.00 ± 2.04 | 52.81 ± 36.05 | 97.50 ± 2.84 |
| Position repair then LoRA r4 n5 | 95.32 ± 2.21 | 87.19 ± 4.87 | 92.81 ± 3.76 | 65.31 ± 30.51 | 98.75 ± 2.04 |
| Projection initialized LoRA n5 | 93.75 ± 2.47 | 89.37 ± 2.04 | 95.63 ± 1.31 | 49.06 ± 34.62 | 98.44 ± 1.56 |
| Gradient SVD initialization n5 | 95.94 ± 2.37 | 88.12 ± 3.05 | 80.62 ± 10.10 | 1.56 ± 3.49 | 87.81 ± 10.79 |
| Live LLM first candidate n3 | 63.02 ± 0.90 | 72.40 ± 1.81 | 74.48 ± 0.90 | 66.14 ± 0.91 | 54.17 ± 38.82 |
| Scripted single candidate n5 | 62.81 ± 1.31 | 71.88 ± 1.57 | 74.06 ± 0.85 | 65.31 ± 2.57 | 62.50 ± 29.73 |
| Live LLM feedback n3 | 63.02 ± 0.90 | 71.88 ± 1.57 | 74.48 ± 0.90 | 65.62 ± 1.57 | 76.56 ± 1.56 |
| Scripted five candidates n5 | 64.69 ± 2.61 | 73.75 ± 2.56 | 74.06 ± 0.85 | 65.31 ± 2.57 | 75.62 ± 1.78 |
| Scripted recipe selector n5 | 63.75 ± 1.31 | 71.88 ± 1.57 | 73.44 ± 1.10 | 65.31 ± 2.57 | 75.31 ± 1.31 |
| Scripted blended operator n5 | 62.19 ± 1.31 | 75.31 ± 1.71 | 74.06 ± 0.85 | 4.06 ± 4.08 | 75.62 ± 2.37 |
| Live LLM repair then LoRA n3 | 96.36 ± 0.91 | 84.37 ± 4.13 | 95.32 ± 2.71 | 71.35 ± 12.14 | 100.00 ± 0.00 |
| Scripted repair then LoRA n5 | 95.94 ± 1.40 | 86.88 ± 2.37 | 92.81 ± 3.76 | 57.50 ± 32.29 | 98.44 ± 2.21 |
| Random search (5 evals) n5 | 68.44 ± 3.39 | 70.00 ± 6.38 | 29.69 ± 20.76 | 0.00 ± 0.00 | 1.56 ± 1.56 |
| Bayesian black-box search (5 evals) n5 | 64.69 ± 3.04 | 73.44 ± 2.47 | 74.06 ± 0.85 | 0.00 ± 0.00 | 75.94 ± 1.40 |

*Table 16. Expanded seed coverage for Test B. Non live methods use five fault seeds, and live methods use three. These mixed seed summaries supplement the matched three-seed comparisons in the main text. The original checkpoint is the same reference in every fault trial.*

| Method | BERT | PubMed BERT | SmolLM2 | Pythia | Qwen2.5 |
|---|---|---|---|---|---|
| Original checkpoint n5 | 53.12 ± 0.00 | 64.06 ± 0.00 | 76.56 ± 0.00 | 14.06 ± 0.00 | 45.31 ± 0.00 |
| Damaged checkpoint n5 | 15.31 ± 4.33 | 46.25 ± 3.60 | 0.00 ± 0.00 | 0.00 ± 0.00 | 0.00 ± 0.00 |
| Repetition penalty n5 | 0.62 ± 0.85 | 3.12 ± 1.91 | 0.00 ± 0.00 | 0.00 ± 0.00 | 0.00 ± 0.00 |
| Oracle inverse n5 | 53.12 ± 0.00 | 64.06 ± 0.00 | 76.56 ± 0.00 | 14.06 ± 0.00 | 45.31 ± 0.00 |
| Position projection n5 | 44.69 ± 2.84 | 53.75 ± 6.41 | 75.62 ± 1.78 | 14.06 ± 1.91 | 35.62 ± 19.93 |
| Gated spectral projection n5 | 49.69 ± 1.31 | 63.75 ± 1.71 | 75.62 ± 1.78 | 14.06 ± 1.91 | 35.62 ± 19.93 |
| Dense position projection n5 | 44.37 ± 2.84 | 53.44 ± 3.73 | 73.13 ± 1.71 | 15.94 ± 1.31 | 43.75 ± 0.00 |
| Rollout diagnostic projection n5 | 44.69 ± 2.84 | 53.75 ± 6.41 | 75.62 ± 1.78 | 14.06 ± 1.91 | 35.62 ± 19.93 |
| Weight SVD projection n5 | 50.00 ± 2.47 | 64.06 ± 3.99 | 75.94 ± 1.78 | 0.00 ± 0.00 | 45.00 ± 0.70 |
| Context damping n5 | 47.50 ± 3.42 | 55.00 ± 6.09 | 74.69 ± 2.31 | 0.00 ± 0.00 | 21.88 ± 13.49 |
| LoRA n5 | 79.69 ± 6.15 | 78.75 ± 4.50 | 65.63 ± 21.62 | 0.00 ± 0.00 | 80.00 ± 17.73 |
| DoRA style n5 | 79.06 ± 5.80 | 80.94 ± 5.11 | 66.88 ± 20.36 | 0.00 ± 0.00 | 80.94 ± 14.88 |
| PiSSA style n5 | 75.31 ± 9.27 | 77.19 ± 8.39 | 78.13 ± 10.37 | 0.00 ± 0.00 | 90.31 ± 3.01 |
| LoReFT style n5 | 76.87 ± 7.11 | 80.94 ± 4.05 | 65.00 ± 14.56 | 0.00 ± 0.00 | 85.00 ± 4.64 |
| Support masked LoRA n5 | 66.25 ± 5.91 | 78.75 ± 5.70 | 69.69 ± 4.37 | 25.00 ± 9.04 | 87.50 ± 4.69 |
| Position repair then LoRA r3 n5 | 76.56 ± 10.99 | 79.06 ± 2.84 | 91.56 ± 2.62 | 35.00 ± 29.82 | 85.62 ± 4.86 |
| Position repair then LoRA r4 n5 | 75.94 ± 9.48 | 80.00 ± 4.99 | 91.25 ± 1.78 | 56.56 ± 23.06 | 88.13 ± 5.70 |
| Projection initialized LoRA n5 | 81.87 ± 6.21 | 83.13 ± 6.57 | 93.44 ± 2.04 | 30.00 ± 30.63 | 91.88 ± 3.21 |
| Gradient SVD initialization n5 | 80.00 ± 5.89 | 85.63 ± 2.79 | 60.00 ± 22.63 | 0.62 ± 1.40 | 77.50 ± 15.25 |
| Live LLM first candidate n3 | 44.79 ± 3.25 | 54.69 ± 5.42 | 74.48 ± 0.90 | 14.58 ± 2.39 | 29.69 ± 25.72 |
| Scripted single candidate n5 | 44.69 ± 2.84 | 53.75 ± 6.41 | 75.62 ± 1.78 | 14.06 ± 1.91 | 35.62 ± 19.93 |
| Live LLM feedback n3 | 50.00 ± 1.56 | 53.64 ± 3.61 | 75.00 ± 1.56 | 14.58 ± 2.39 | 44.79 ± 0.90 |
| Scripted five candidates n5 | 48.12 ± 4.61 | 58.12 ± 11.07 | 76.56 ± 1.91 | 14.06 ± 1.91 | 41.25 ± 7.38 |
| Scripted recipe selector n5 | 45.94 ± 7.21 | 53.75 ± 6.59 | 75.00 ± 2.21 | 14.06 ± 1.91 | 40.62 ± 7.97 |
| Scripted blended operator n5 | 49.38 ± 0.85 | 64.06 ± 0.00 | 74.38 ± 2.83 | 0.00 ± 0.00 | 36.56 ± 7.54 |
| Live LLM repair then LoRA n3 | 80.21 ± 6.31 | 83.33 ± 2.39 | 89.06 ± 2.71 | 53.12 ± 14.91 | 84.90 ± 3.25 |
| Scripted repair then LoRA n5 | 80.31 ± 5.59 | 80.31 ± 4.64 | 90.31 ± 2.04 | 39.69 ± 32.84 | 86.25 ± 3.00 |
| Random search (5 evals) n5 | 31.25 ± 14.32 | 42.81 ± 7.86 | 10.00 ± 8.16 | 0.00 ± 0.00 | 0.00 ± 0.00 |
| Bayesian black-box search (5 evals) n5 | 39.06 ± 11.27 | 55.94 ± 10.33 | 75.94 ± 1.78 | 0.00 ± 0.00 | 45.00 ± 0.70 |

# B Recorded live sessions

*Table 17. Manifest for the 20 recorded live sessions. The candidate choices are session identifiers. Three BERT conversation identifiers are absent; all editor model fields use a generic role identifier.*

| Target | Seed | Fault | Choice | Dev calls | A hits first / final | Conversation ID |
|---|---|---|---|---|---|---|
| BERT | 101 | Core | 2 | 3 | 40 / 40 | absent |
| BERT | 202 | Core | 2 | 3 | 41 / 40 | absent |
| BERT | 303 | Core | 2 | 3 | 40 / 41 | absent |
| PubMed BERT | 101 | Core | 1 | 3 | 45 / 45 | yes |
| PubMed BERT | 202 | Core | 2 | 3 | 47 / 46 | yes |
| PubMed BERT | 303 | Core | 1 | 3 | 47 / 47 | yes |
| Pythia | 101 | Core | 1 | 3 | 43 / 43 | yes |
| Pythia | 202 | Core | 1 | 3 | 42 / 42 | yes |
| Pythia | 303 | Core | 3 | 3 | 42 / 41 | yes |
| Qwen2.5 | 101 | Core | 2 | 3 | 6 / 49 | yes |
| Qwen2.5 | 202 | Core | 1 | 3 | 50 / 50 | yes |
| Qwen2.5 | 303 | Core | 1 | 3 | 48 / 48 | yes |
| SmolLM2 | 101 | Random | 2 | 4 | 48 / 48 | yes |
| SmolLM2 | 101 | K32 | 3 | 4 | 49 / 47 | yes |
| SmolLM2 | 101 | K256 | 2 | 3 | 10 / 47 | yes |
| SmolLM2 | 101 | Rank4 | 3 | 4 | 49 / 47 | yes |
| SmolLM2 | 101 | Layers4 | 3 | 3 | 12 / 49 | yes |
| SmolLM2 | 101 | Core | 2 | 3 | 48 / 48 | yes |
| SmolLM2 | 202 | Core | 2 | 3 | 47 / 47 | yes |
| SmolLM2 | 303 | Core | 2 | 3 | 48 / 48 | yes |

Candidate identifiers refer to choices within sessions. Finalization or post-hoc evaluation may re-materialize a proposal under another candidate identifier without another development evaluation. The development budget counts explicit development calls, not every diagnostic forward, SVD calculation, or sealed test call. State hashes are truncated checksums, not digital signatures authenticating a provider model or a complete conversation.

# C Statistical interpretation

*Table 18. Paired exact-count statistics reproduced from raw checkpoint seed hits across all five targets, including comparisons with the matched-budget Bayesian black-box optimizer.*

| Model | Comparison | Seeds | Difference pp | Wilcoxon p | Holm p |
|---|---|---|---|---|---|
| BERT | Feedback vs first candidate | 3 | +0.00 | 1.0000 | 1.0000 |
| BERT | Live LLM feedback vs Bayesian black-box | 3 | -1.56 | 0.5000 | 1.0000 |

| Model | Comparison | Seeds | Difference pp | Wilcoxon p | Holm p |
|---|---|---|---|---|---|
| BERT | Bayesian black-box vs Random search | 5 | -3.75 | 0.1875 | 1.0000 |
| BERT | Live LLM repair plus LoRA vs LoRA | 3 | -1.04 | 0.7500 | 1.0000 |
| PubMed BERT | Feedback vs first candidate | 3 | -0.52 | 1.0000 | 1.0000 |
| PubMed BERT | Live LLM feedback vs Bayesian black-box | 3 | -2.60 | 0.5000 | 1.0000 |
| PubMed BERT | Bayesian black-box vs Random search | 5 | +3.44 | 0.5000 | 1.0000 |
| PubMed BERT | Live LLM repair plus LoRA vs LoRA | 3 | +0.52 | 1.0000 | 1.0000 |
| SmolLM2 | Feedback vs first candidate | 3 | +0.00 | 1.0000 | 1.0000 |
| SmolLM2 | Live LLM feedback vs Bayesian black-box | 3 | +0.00 | 1.0000 | 1.0000 |
| SmolLM2 | Bayesian black-box vs Random search | 5 | +44.37 | 0.0625 | 0.5625 |
| SmolLM2 | Live LLM repair plus LoRA vs LoRA | 3 | +8.33 | 0.2500 | 1.0000 |
| Pythia | Feedback vs first candidate | 3 | -0.52 | 1.0000 | 1.0000 |
| Pythia | Live LLM feedback vs Bayesian black-box | 3 | +65.62 | 0.2500 | 1.0000 |
| Pythia | Bayesian black-box vs Random search | 5 | +0.00 | 1.0000 | 1.0000 |
| Pythia | Live LLM repair plus LoRA vs LoRA | 3 | +71.35 | 0.2500 | 1.0000 |
| Qwen2.5 | Feedback vs first candidate | 3 | +22.40 | 1.0000 | 1.0000 |
| Qwen2.5 | Live LLM feedback vs Bayesian black-box | 3 | +0.00 | 1.0000 | 1.0000 |
| Qwen2.5 | Bayesian black-box vs Random search | 5 | +74.38 | 0.0625 | 0.5625 |
| Qwen2.5 | Live LLM repair plus LoRA vs LoRA | 3 | +10.42 | 0.2500 | 1.0000 |

All forty-five saved pairwise comparisons across the five checkpoint-backed targets reproduce exactly from the raw seed hit counts under per-target Holm-Bonferroni correction. With three or five fault seeds, nonparametric exact tests have bounded power; therefore, effect magnitudes and geometric restoration metrics should be interpreted alongside p-values.

Primary task means using integer hit counts where available. The sample SD measures the variation across fault seeds. We avoid normalized-recovery ratios for encoders when original and damaged accuracies are almost equal because dividing by that small difference is unstable. Raw task scores and weight errors are more interpretable in these cases.

# D Implementation checks and reproducibility

An independent check recomputed the best possible full-first-clause lookup bound from exact clause strings and confirmed 50% in every split. Training, validation, and Test A share some distractor strings despite disjoint target strings and full prompts. Test B changes syntax and distractors but reuses Test A target entities. These relationships are part of the data definition.

The current synthetic logging-invariance gate initially fails because its mock target lacks the output-embedding interface expected by the production evaluator. Adding that method only to an isolated test fixture yields identical item vectors with logging on and off for batch sizes 1, 8, 16, and 64:25% completion accuracy, 75% repetition, and 100% first-token accuracy in the counterexample. This validates the intended evaluation property without silently modifying production code or describing the unmodified gate as passing.

The existing Pythia CPU gate passes sequential and initialized-hybrid step-zero checks with zero maximum-logit difference and restores the original state checksum. The production step-zero assertion is looser, allowing a difference below 0.05; saved Pythia differences can be around 0.0007. The gate does not execute official DoRA forward or gradient parity. The live-repair replay helper reconstructs a proposal but does not compare its returned state hash with the committed transcript state before adaptation. Therefore, portable replay requires saved deltas and an explicit equality or numerical-tolerance check.

The preservation corpus has thirty-two sentences. Causal PPL scores valid adjacent-token transitions; the encoder loss masks one selected token per sentence. These are not interchangeable language modeling metrics. Natural completion uses first-word scoring and a short repetition detector; therefore, acceptable alternative phrasing and longer failures can be missed. The saved natural item vectors support their counts, whereas complete token histories are not retained for every condition.

All five targets (BERT-Base, PubMedBERT, SmolLM2, Pythia, and Qwen2.5) have complete raw checkpoint records across all five fault seeds, all eight fault family sweeps, all three training horizons (25, 50, and 100 steps), the matched-budget Bayesian black-box control, and the vingt recorded live editor sessions.